\documentclass{article}

\usepackage{microtype}
\usepackage{graphicx}
\usepackage{subcaption}
\usepackage{booktabs}
\usepackage[colorlinks=true,linkcolor=blue!70!black,citecolor=green!50!black,urlcolor=blue!70!black]{hyperref}

\usepackage[accepted]{icml2026}

\makeatletter
\renewcommand{\ICML@appearing}{}
\makeatother

\usepackage{amsmath}
\usepackage{amssymb}
\usepackage{mathtools}
\usepackage{amsthm}
\usepackage{bm}

\usepackage[capitalize,noabbrev]{cleveref}

\usepackage{enumitem}

\usepackage{xurl}

\usepackage{multirow}
\usepackage{colortbl}

\usepackage{amsmath,amsfonts,bm}

\def\eqref#1{equation~\ref{#1}}

\def\1{\bm{1}}

\DeclareMathAlphabet{\mathsfit}{\encodingdefault}{\sfdefault}{m}{sl}
\SetMathAlphabet{\mathsfit}{bold}{\encodingdefault}{\sfdefault}{bx}{n}

\definecolor{bestrow}{rgb}{0.9, 1.0, 0.9}
\definecolor{secondrow}{rgb}{0.95, 0.95, 1.0}

\def\github{\raisebox{-1.5pt}{\includegraphics[height=1.05em]{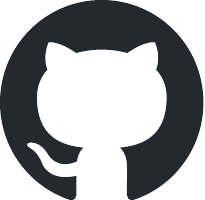}}\hspace{0.3em}}
\newcommand{\ghlink}{https://github.com/seonglae/confidence-manifold}

\icmltitlerunning{The Confidence Manifold}

\AtBeginDocument{%
  \hypersetup{citecolor=green!50!black}%
}

\begin{document}

\twocolumn[
\icmltitle{The Confidence Manifold: \\
Geometric Structure of Correctness Representations in Language Models}

\icmlsetsymbol{equal}{*}

\begin{icmlauthorlist}
\icmlauthor{Seonglae Cho}{holistic,ucl}
\icmlauthor{Zekun Wu}{holistic,ucl}
\icmlauthor{Kleyton Da Costa}{holistic}
\icmlauthor{Adriano Koshiyama}{holistic,ucl}
\end{icmlauthorlist}

\icmlaffiliation{holistic}{Holistic AI}
\icmlaffiliation{ucl}{University College London}
\icmlcorrespondingauthor{Seonglae Cho}{seonglae.cho@holisticai.com}

\icmlkeywords{Correctness Detection, Interpretability, Language Models, Confidence Estimation, Geometric Analysis}

\vskip 0.2in

\begin{center}
\github \url{\ghlink}
\end{center}

\vskip 0.1in
]

\printAffiliationsAndNotice{}

\begin{abstract}
When a language model asserts that ``the capital of Australia is Sydney,'' does it know this is wrong? Models assert misconceptions with the same fluency as facts, so the question cannot be answered from output uncertainty. Truth-related signals are known to exist in the residual stream, but not their geometry: how many dimensions carry the signal, how simple a detector can be, and whether it transfers. We characterize this geometry across 11 models (124M--14B) and test it causally with activation steering, concept erasure, and distributed alignment search. The structure is simple: two class centroids in a 2--8 dimensional subspace match a trained linear probe, and 25 labeled examples recover 90\% of full-data AUC on GPT-2. Steering shifts hallucination rates by 9.1 points on six models, erasure drops detection to chance, and distributed alignment search, the only method that bounds rank, localizes at most five causal dimensions. The internal advantage is regime-specific: probes far outperform P(True) and semantic entropy on adversarial misconceptions but tie on standard QA. Single-dataset probes transfer near-randomly until joint multi-dataset training restores 0.73--0.91 AUC. That centroid distance matches probe performance indicates class separation is a mean shift, making detection geometric rather than learned.

\end{abstract}

\section{Introduction}
\label{sec:intro}

When a language model states that ``the capital of Australia is Sydney,'' does it internally register that the claim is false? Language models assert falsehoods with the same fluency as facts \citep{rawte-etal-2023-troubling, ji-etal-2025-language-models}, which makes them most dangerous exactly when their confidence and their errors are indistinguishable. A growing body of work suggests the model often does register the error: truth-related signals are linearly present in the residual stream \citep{azaria2023internal, burns2023discovering, marks2024geometry, orgad2025llmsknow}, and a low-dimensional subspace can separate true from false statements \citep{burger2024truth}.

This prior work establishes that such a subspace \emph{exists}; it leaves open how to \emph{use} it. Three questions matter for a detector: how simple the decision rule can be before the signal is lost, how many dimensions the correctness direction causally occupies rather than merely correlates with, and whether a direction learned on one task transfers to another. Difference-of-means probing is reported to work \citep{marks2024geometry}, but without characterizing why two means suffice or when the direction generalizes.

We answer these questions by characterizing the \emph{class-conditional discriminative subspace} for correctness across 11 models from five families (124M--14B), and validating it with three complementary causal methods: activation steering, concept erasure \citep{ravfogel2022linear}, and distributed alignment search (DAS) \citep{wu2023interpretability}.\footnote{All AUC values are area under the ROC curve.}

\begin{itemize}
\item \textbf{Two centroids suffice.} Distance to two class centroids in a supervised 2--8D subspace matches a trained probe within $\pm0.02$ AUC and beats it at larger label budgets, while multi-prototype variants degrade: the per-class distributions are unimodal, so one mean shift is the decision rule.
\item \textbf{The signal is causally low-rank.} Steering along the direction shifts hallucination rates (9.1pp on Qwen2-7B) and erasing the subspace collapses detection to chance, establishing that the direction is sufficient and necessary; DAS independently bounds the causal rank at five dimensions.
\item \textbf{Cross-domain transfer needs joint training.} Single-dataset probes transfer near-randomly (0.39--0.66 AUC), but joint multi-dataset training restores 0.73--0.91 AUC on instruction-tuned models 1B and larger.
\end{itemize}

\begin{figure}[t]
\centering
\includegraphics[width=\linewidth]{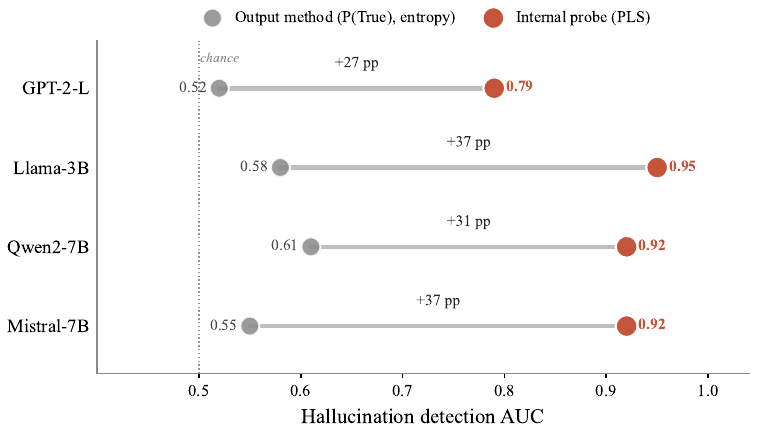}
\caption{Internal probes beat output-based uncertainty (P(True), semantic entropy) on TruthfulQA's adversarial misconception questions; the advantage closes on standard QA (\S\ref{sec:results}).}
\label{fig:compression_cascade}
\end{figure}

We present these as properties of correctness geometry, not a new detector: the low-rank separability of truth is established \citep{marks2024geometry, burger2024truth}, whereas centroid \emph{sufficiency}, a \emph{bounded} causal rank, and \emph{recipe-dependent} transfer are not. The advantage over output-based uncertainty is regime-specific (Figure~\ref{fig:compression_cascade}): internal probes win on adversarial misconception data but only tie on standard QA, so the consistent claim is centroid sufficiency, not internal superiority. Our probes are trained and evaluated teacher-forced, as is standard in this line of work \citep{marks2024geometry, burns2023discovering, orgad2025llmsknow}; we characterize where the correctness signal lives and how it is structured, and leave deployment to free-form generation to future work.

\section{Background}
\label{sec:background}

\textbf{Linear Representation Hypothesis.}
Neural networks encode concepts linearly \citep{mikolov2013linguistic, park2024the}: $\mathbf{w}_c$ enables extraction and intervention \citep{nanda2023emergent, marks2024geometry}.
This extends to truth \citep{burns2023discovering, marks2024geometry}, internal states \citep{azaria2023internal, su2024unsupervised, sriramanan2024llmcheck}, semantic alignment \citep{saglam2025separable}, intrinsic-dimension correlation with truthfulness \citep{yin2024characterizing}, and confidence regulation \citep{stolfo2024confidence}; LLMs encode more truthfulness than they express \citep{orgad2025llmsknow}.
Prior work establishes \emph{existence} and individual properties (linear extractability \citep{marks2024geometry}, pooled local intrinsic dimension \citep{yin2024characterizing}); we characterize the \emph{class-conditional discriminative subspace} jointly across 11 models: dimensionality (PLS), decision boundary (mean shift), and causal structure, where steering tests efficacy, concept erasure \citep{ravfogel2022linear} tests necessity, and DAS \citep{wu2023interpretability} bounds rank.

\textbf{Uncertainty vs.\ Correctness.}
Token entropy $H(p) = -\sum_i p_i \log p_i$ conflates linguistic and epistemic uncertainty, and pre-trained transformers are systematically overconfident to begin with \citep{desai2020calibration}.
Semantic entropy \citep{kuhn2023semantic, farquhar2024semantic} addresses this by clustering $N$ generations by meaning via NLI, computing entropy over clusters.
While effective for uncertainty estimation, it requires $N$ forward passes plus NLI inference.
Calibration evolves across layers with a low-dimensional direction in the residual stream \citep{joshi2025calibration}, but distributional certainty alone is insufficient: models can be \emph{confidently wrong} on TruthfulQA's misconception-laden questions.
Semantic entropy probes \citep{kossen2024semantic} predict SE efficiently but inherit its limitation of measuring uncertainty rather than correctness.

\textbf{Activation Steering.}
Inference-time intervention modifies activations: $\mathbf{h}' = \mathbf{h} + \alpha \mathbf{w}$ \citep{li2023inferencetime, turner2023steeringlanguagemodelsactivation}.
Contrastive activation addition \citep{rimsky-etal-2024-steering} and representation engineering \citep{zou2023representation} demonstrate behavioral control for honesty and safety.
Adaptive steering \citep{wang2024adaptive} adjusts intervention strength per-sample based on predicted uncertainty.
Causal mediation analysis \citep{vig2020causal} and causal abstraction \citep{geiger2024causal} ask \emph{where} in the network a behavior is computed; our question is instead \emph{how many dimensions} carry it, so we use steering, erasure, and DAS for causal validation of the subspace rather than for localization.

\textbf{Problem Setting.}
Given question $q$ and response $s$, predict factual correctness $y \in \{0,1\}$ from hidden state $\mathbf{h}^{(\ell)}$ at layer $\ell$.
Goals: (1) single-pass inference, (2) improvement over entropy baselines, (3) causal validation via intervention.

\section{Method}
\label{sec:method}

We extract, characterize, and validate the correctness representation through contrastive data construction, direction learning, geometric analysis, and causal validation.\footnote{Three terms are distinguished as follows. \textbf{Embedding manifold} (Levina--Bickel MLE on pooled activations, \S\ref{sec:results}): the surface residual-stream activations live on. \textbf{Discriminative subspace} (supervised PLS, \S\ref{sec:results}): the linear subspace within which correctness is most cleanly separated. \textbf{Causal subspace} (DAS, \S\ref{sec:results}): the dimensions where intervention flips predicted correctness. ``Confidence'' denotes the model's internal representation of correctness, not output-level uncertainty.}

\textbf{Definitions.} For input $x$ and answer $y$, let $c(x, y) \in \{0, 1\}$ be the binary \emph{correctness} label (1 if $y$ is the gold answer, 0 otherwise) and $h_\ell(x, y) \in \mathbb{R}^d$ the residual-stream activation at layer $\ell$. We distinguish two scalar predictors of $c$:
\emph{output uncertainty} $u_\text{out}$, derived from output logits without supervision (e.g.\ NLL $-\log p_\theta(y \mid x)$, P(True), token entropy, semantic entropy);
and \emph{internal confidence} $u_\text{int}(x, y) = f\!\big(h_\ell(x, y)\big)$, a learned scalar derived from a hidden state via a function $f$ supervised on $c$.
Correctness is the target; uncertainty and confidence are two scalar predictors that differ only in whether they read the output distribution or the residual stream.

\textbf{Formulation.} We hypothesize that $f$ admits a low-rank linear factorization through a subspace $\mathcal{M} \subseteq \mathbb{R}^d$ of dimension $k \ll d$: (1) projection onto $\mathcal{M}$ predicts $c$, (2) intervention along $\mathcal{M}$ causally affects predicted correctness, (3) $\mathcal{M}$ has consistent structure across layers and architectures.

\subsection{Data Construction and Direction Learning}

\textbf{Contrastive pairs.} TruthfulQA provides paired correct/incorrect answers per question \citep{lin2022truthfulqa}, yielding $\{(\mathbf{h}^+_i, \mathbf{h}^-_i)\}_{i=1}^N$ controlling for topic and difficulty.
Each pair shares the same question stem, isolating the correctness signal from content variation.
We extract hidden states from the last token position \citep{gurnee2023finding, belinkov2022probing} across all $L$ layers.

\textbf{Probe training.} Logistic regression with $L_2$ regularization ($C$=0.1): $p(y\!=\!1 | \mathbf{h}) = \sigma(\mathbf{w}^\top \mathbf{h} + b)$, trained with cross-entropy loss plus $L_2$ penalty.
The learned $\mathbf{w}^{(\ell)}$ is the correctness direction at layer $\ell$.
Probe labels are TruthfulQA's binary correct/incorrect annotations, not derived from model entropy or uncertainty estimates; probes trained on entropy proxies fail to separate correct from incorrect samples (\S\ref{sec:results}).

\subsection{Geometric Analysis}

We distinguish two notions of dimensionality.

\textbf{Intrinsic dimension (representation geometry).} The Levina--Bickel MLE estimator \citep{levina2004maximum} measures the dimensionality of the data manifold itself: for each point $\mathbf{x}_i$ with $k$-nearest neighbors at distances $T_1 < \cdots < T_k$,
\begin{equation}
\hat{d}_k(\mathbf{x}_i) = \left( \frac{1}{k-1} \sum_{j=1}^{k-1} \log \frac{T_k}{T_j} \right)^{-1}
\end{equation}
averaged over samples and $k \in [5, 20]$.
We pool correct and incorrect activations to estimate the intrinsic dimension of the representation manifold at each layer, independent of any classification task.

\textbf{Discriminative dimension (classification geometry).} Partial Least Squares (PLS) regression projects activations onto directions maximizing covariance with labels.
Sweeping the number of PLS components measures how many dimensions are useful for classification (Table~\ref{tab:dimension_sweep}).

\subsection{Causal Validation}

We establish causality with three methods that answer different questions: activation steering (is the direction causal?), distributed alignment search (how few dimensions carry the effect?), and concept erasure (is the subspace necessary?).

\textbf{Activation steering.} Inference-time intervention on the residual stream \citep{li2023inferencetime, turner2023steeringlanguagemodelsactivation}:
\begin{equation}
\mathbf{h}'^{(\ell^*)} = \mathbf{h}^{(\ell^*)} + \alpha \cdot \hat{\mathbf{w}}^{(\ell^*)}.
\end{equation}
Unlike ITI \citep{li2023inferencetime}, which intervenes at attention-head outputs, we steer the full residual stream, consistent with the discriminative subspace being identified there.
Controls are a random direction $\mathbf{r} \sim \mathcal{N}(\mathbf{0}, \mathbf{I})$ and its component orthogonal to $\hat{\mathbf{w}}$, both normalized.

\textbf{Distributed Alignment Search (DAS).} Following \citet{wu2023interpretability, geiger2024causal}, we learn a rotation $\mathbf{R} \in \mathbb{R}^{d \times k}$ with an interchange-intervention objective: swapping the $k$-dimensional projected representation between correct and incorrect samples should flip the prediction.
DAS is a \emph{causal} operationalization of dimensionality, complementing PLS's statistical one, and is the only one of the three methods that bounds the rank of the causal subspace.

\textbf{Concept erasure (RLACE).} Linear Adversarial Concept Erasure \citep{ravfogel2022linear} removes the correctness signal by projecting activations onto the nullspace of the discriminative direction.
If the subspace is necessary, erasure should reduce AUC to chance ($\sim$0.50).
We also apply Iterative Nullspace Projection (INLP; \citealp{ravfogel2020null}) as a complementary iterative erasure method.

\section{Experiments}
\label{sec:experiments}

\textbf{Datasets.} We evaluate on five QA datasets: TruthfulQA \citep{lin2022truthfulqa} (817 questions with paired correct/incorrect answers), SciQ \citep{welbl2017crowdsourcing}, CommonsenseQA \citep{talmor2019commonsenseqa}, FEVER \citep{thorne2018fever}, and HaluEval \citep{li2023halueval}.
The discriminative-subspace dim sweep is run natively on the first four; cross-dataset transfer is evaluated on all five.
We use 80/20 stratified splits with GroupKFold (5 folds, grouped by question) to prevent paraphrase leakage.

\textbf{Models (11 across 5 families, 124M--14B).} GPT-2 family \citep{radford2019language} (124M/355M/774M); Qwen2 \citep{yang2024qwen2} (1.5B/7B), Qwen2.5-14B, Mistral-7B \citep{jiang2023mistral7b}, Llama-3.2 \citep{grattafiori2024llama3herdmodels} (1B/3B), Llama-3.1-8B, and Gemma-2-2B \citep{gemmateam2024gemma2improvingopen}.

\textbf{Baselines.} \emph{Output-based:} P(True) \citep{kadavath2022language}, NLL, token entropy, verbalized confidence, semantic entropy \citep{farquhar2024semantic}.
\emph{Unsupervised:} CCS \citep{burns2023discovering}, L2 norm, reconstruction error, LOF, cluster uncertainty.

\textbf{Probing protocol.} Last-token residual stream activations \citep{gurnee2023finding, belinkov2022probing}; logistic regression with $L_2$ regularization ($C{=}0.1$); GroupKFold (5-fold) ensures no question in both train and test.
Preprocessing (standardization, PLS) is fit on training folds only.

\textbf{Validation.} Nested CV (outer 5-fold for evaluation, inner 3-fold for layer/dimension selection) shows negligible bias: Qwen2-7B $+0.005$, GPT-2-Large $-0.026$ (Appendix~\ref{app:nested_cv}).
Length-balanced evaluation, length-residualized probing, and surface-feature correlations rule out trivial confounds.
A contamination check (200 generated answers vs.\ ground truth) yields 0/200 exact matches and 13/200 substring matches.
All experiments run on NVIDIA GB10 and RTX-class GPUs under PyTorch 2.9 with HuggingFace Transformers and scikit-learn; seeds and implementation details are in Appendix~\ref{app:reproducibility}, and code is available at the repository linked on the title page.

\section{Results}
\label{sec:results}

A correctness probe reaches 0.80--0.97 AUC across the 11 models (Table~\ref{tab:main_results}), reproducing the established finding that correctness is linearly decodable \citep{marks2024geometry, orgad2025llmsknow}.
What follows asks what structure that decodability has: how few dimensions it needs, how simple the decision rule can be, which causal claims the interventions license, and when the direction transfers.

\begin{table}[t]
\centering
{\small
\begin{tabular}{@{}lcrcc@{}}
\toprule
\textbf{Model} & \textbf{Size} & \textbf{Layer} & \textbf{Depth} & \textbf{AUC} \\
\midrule
GPT-2          & 124M & L11/12 & 100\% & 0.80$_{\pm.04}$ \\
GPT-2-Med      & 355M & L23/24 & 100\% & 0.84$_{\pm.02}$ \\
GPT-2-Large    & 774M & L35/36 & 100\% & 0.84$_{\pm.02}$ \\
Llama-3.2-1B   & 1B   & L8/16  & 56\%  & 0.93$_{\pm.02}$ \\
Qwen2-1.5B     & 1.5B & L16/28 & 61\%  & 0.91$_{\pm.03}$ \\
Gemma-2-2B     & 2B   & L15/26 & 62\%  & 0.93$_{\pm.01}$ \\
Llama-3.2-3B   & 3B   & L12/28 & 46\%  & \textbf{0.97}$_{\pm.01}$ \\
Qwen2-7B       & 7B   & L20/28 & 75\%  & 0.94$_{\pm.02}$ \\
Mistral-7B     & 7B   & L23/32 & 75\%  & 0.92$_{\pm.02}$ \\
Llama-3.1-8B   & 8B   & L16/32 & 53\%  & 0.92$_{\pm.03}$ \\
Qwen2.5-14B    & 14B  & L32/48 & 69\%  & 0.95$_{\pm.02}$ \\
\bottomrule
\end{tabular}
}
\caption{Correctness detection across 11 models (TruthfulQA, GroupKFold AUC). Layer = optimal layer / total layers.}
\label{tab:main_results}
\end{table}

\subsection{The Correctness Signal is 2--8 Dimensional}
\label{sec:res_dim}

\definecolor{peakcell}{rgb}{0.83,0.91,1.0}
\definecolor{degradecell}{rgb}{0.97,0.91,0.91}

\begin{table*}[t]
\centering
{\small
\setlength{\tabcolsep}{4pt}
\begin{tabular}{@{}lrrrrrrrrc@{}}
\toprule
\textbf{Model} & \textbf{1D} & \textbf{2D} & \textbf{3D} & \textbf{4D} & \textbf{5D} & \textbf{8D} & \textbf{16D} & \textbf{32D} & \textbf{Peak} \\
\midrule
GPT-2          & 0.718 & 0.750 & \cellcolor{peakcell}\textbf{0.758} & 0.753 & 0.746 & 0.720 & \cellcolor{degradecell}0.672 & \cellcolor{degradecell}0.621 & 3D \\
GPT-2-Med      & 0.747 & 0.781 & \cellcolor{peakcell}\textbf{0.790} & 0.784 & 0.778 & 0.763 & \cellcolor{degradecell}0.723 & \cellcolor{degradecell}0.676 & 3D \\
GPT-2-Large    & 0.750 & 0.784 & \cellcolor{peakcell}\textbf{0.791} & 0.783 & 0.769 & \cellcolor{degradecell}0.741 & \cellcolor{degradecell}0.700 & \cellcolor{degradecell}0.673 & 3D \\
Llama-3.2-1B   & 0.847 & 0.886 & 0.892 & \cellcolor{peakcell}\textbf{0.895} & 0.893 & 0.878 & \cellcolor{degradecell}0.831 & \cellcolor{degradecell}0.813 & 4D \\
Qwen2-1.5B     & 0.814 & 0.848 & 0.858 & 0.863 & 0.864 & \cellcolor{peakcell}\textbf{0.873} & 0.842 & \cellcolor{degradecell}0.810 & 8D \\
Gemma-2-2B     & 0.817 & 0.846 & \cellcolor{peakcell}\textbf{0.861} & 0.860 & 0.856 & 0.830 & \cellcolor{degradecell}0.780 & \cellcolor{degradecell}0.751 & 3D \\
Llama-3.2-3B   & 0.887 & 0.914 & 0.917 & 0.917 & \cellcolor{peakcell}\textbf{0.919} & 0.910 & 0.893 & 0.881 & 5D \\
Qwen2-7B       & \cellcolor{degradecell}0.828 & 0.882 & 0.892 & 0.894 & \cellcolor{peakcell}\textbf{0.896} & 0.890 & 0.876 & \cellcolor{degradecell}0.843 & 5D \\
Mistral-7B     & 0.860 & 0.888 & 0.899 & 0.902 & \cellcolor{peakcell}\textbf{0.902} & 0.899 & 0.876 & \cellcolor{degradecell}0.849 & 5D \\
Llama-3.1-8B   & 0.895 & \cellcolor{peakcell}\textbf{0.931} & 0.929 & 0.929 & 0.923 & 0.910 & 0.900 & 0.890 & 2D \\
Qwen2.5-14B    & 0.927 & \cellcolor{peakcell}\textbf{0.951} & 0.951 & 0.951 & 0.949 & 0.943 & 0.938 & 0.939 & 2D \\
\bottomrule
\end{tabular}
}
\caption{\textbf{All 11 models peak at 2--8 dimensions and degrade beyond.} Adding dimensions hurts: GPT-2 loses 18\% AUC from 3D to 32D, Mistral-7B 5\%, Qwen2.5-14B 1\%. PLS subspace dimension sweep on TruthfulQA, GroupKFold AUC; per-fold standard deviations in Appendix~\ref{app:dimension_sweep_std}. \colorbox{peakcell}{Blue} = peak; \colorbox{degradecell}{red} = $\geq$5pp degradation from peak.}
\label{tab:dimension_sweep}
\end{table*}

We sweep the dimensionality of a supervised PLS subspace, which maximizes covariance with the correctness label, and evaluate a probe inside each subspace (Table~\ref{tab:dimension_sweep}).
Every model peaks between 2 and 8 dimensions and then \emph{degrades}: the GPT-2 family peaks at 3D, larger models at 2--8D, and GPT-2 falls from 0.76 (3D) to 0.62 (32D), an 18\% drop.
Degradation beyond the peak is the informative part.
If correctness were spread across many directions, adding dimensions would be neutral at worst; instead the extra directions carry variance that is uninformative about correctness, so a fixed label budget spends capacity on noise.

\textbf{What the dimensions encode is open.}
We locate the subspace but do not identify what its individual directions represent.
The count falls in the same range as the three confidence mechanisms \citet{stolfo2024confidence} report, but we treat that as a coincidence of magnitude, not evidence of correspondence: a covariance-maximizing basis fixes the subspace only up to rotation within it, so no single PLS component has a semantics to read off.
Naming these directions needs a decomposition with an identifiability guarantee, which we leave open.

\subsection{Two Centroids Suffice}
\label{sec:res_centroid}

Inside that subspace the decision rule collapses to something far simpler than a trained classifier.
All geometric classifiers we tested (Mahalanobis distance, $k$-nearest neighbors, kernel SVM, nearest convex hull, centroid distance) land within $\pm0.02$ AUC of the linear probe across 9 models (Appendix~\ref{app:geometric_methods}), and the optimal boundary is a hyperplane between the two class means:
\begin{equation*}
\text{conf}(x) = \frac{\exp(-\|x - \mu_\text{c}\|)}{\exp(-\|x - \mu_\text{c}\|) + \exp(-\|x - \mu_\text{i}\|)},
\end{equation*}
where $\mu_\text{c}$ and $\mu_\text{i}$ are the correct- and incorrect-class means.
Separation is therefore dominated by a \emph{mean shift} rather than by a difference in class covariances, which is why methods that model covariance buy nothing.

Three checks rule out the obvious alternative explanations.
First, the effect is not manufactured by PLS: difference-of-means in the \emph{raw} residual stream, with no projection, already reaches 0.72--0.91 AUC, and PLS adds only $+0.034$ to $+0.081$ (mean $+0.057$, Appendix~\ref{app:dom_baseline}).
The mean shift lives in the original representation and is merely sharpened by class-conditional projection.
Second, labels are doing real work: unsupervised features plateau at 0.51--0.62 AUC, individual pretrained SAE features explain $<$6\% of variance, and the pooled local intrinsic dimension of \citet{yin2024characterizing} reaches only 0.51--0.58, a $+0.34$ mean gap to the PLS centroid that separates our class-conditional subspace from label-free dimension estimates (Appendices~\ref{app:unsupervised},~\ref{app:sae},~\ref{app:yin_lid}).
Third, extra prototypes hurt: KMeans-2 and KMeans-3 within each class cost $1.3$--$2.4$pp, so the per-class distributions behave as single unimodal clusters and two means are the right model (Appendix~\ref{app:centroid_methods}).

Centroid distance is also the more label-efficient rule.
On GPT-2 it rises from 0.60 AUC at 5 labels to 0.69 at 25 (90\% of the full-data value) and 0.76 at 100, matching or exceeding the trained probe at every budget (Appendix~\ref{app:fewshot}).
A head-to-head sweep on Qwen2-7B over $K \in \{10, 25, 50, 100, 200, 400\}$ shows the same shape: centroid and LR are within $\pm0.5$pp of each other up to $K = 100$, and centroid pulls ahead by $+1.2$--$1.5$pp once $K \geq 200$ (0.907 vs 0.894 at $K = 400$; Appendix~\ref{app:centroid_methods}).
Two means estimated from a few dozen examples are thus not a degraded approximation of a probe; at realistic label budgets they are the better estimator, because a mean is cheaper to estimate than a separating hyperplane.

\subsection{What the Interventions Do and Do Not Show}
\label{sec:res_causal}

Steering, concept erasure, and DAS are often bundled as joint evidence for a low-rank causal subspace.
They are not interchangeable, and we separate them because only one of the three constrains rank.

\begin{figure*}[t]
\centering
\includegraphics[width=0.72\linewidth]{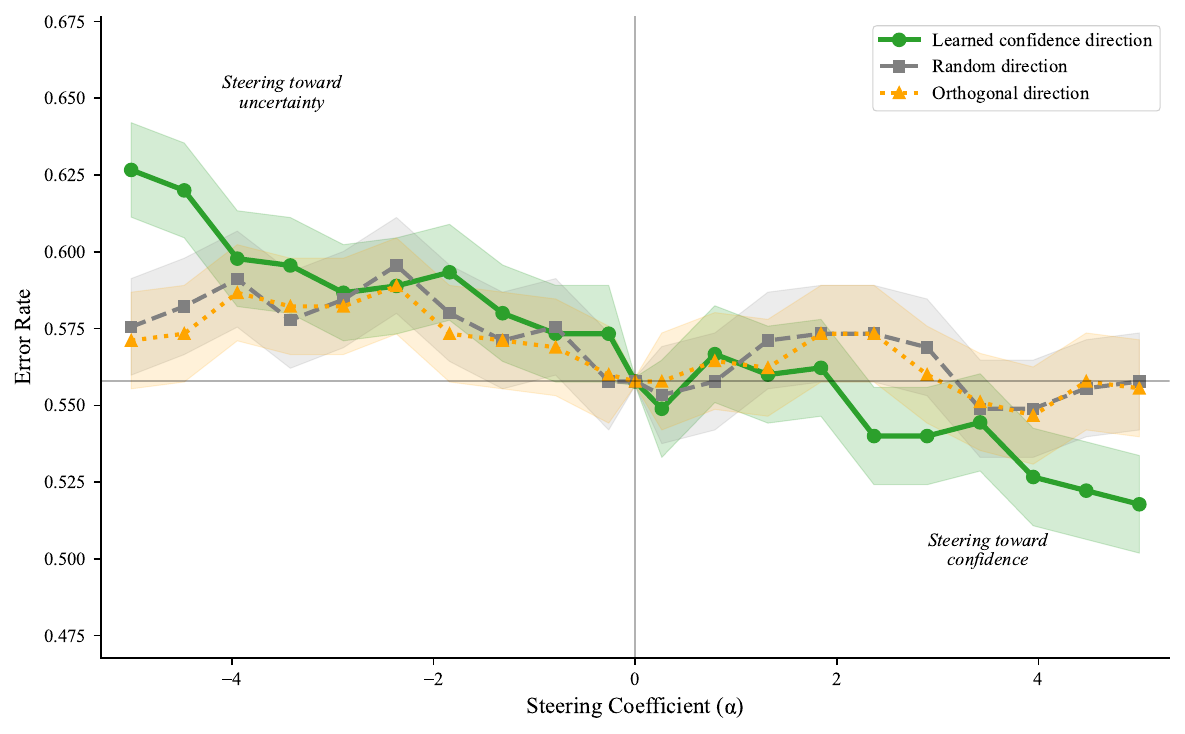}
\caption{\textbf{Steering intervention on Qwen2-7B-Instruct} ($n=617$ held-out TruthfulQA questions). Error rate vs.\ steering coefficient $\alpha \in [-5, 5]$.
Interventions modify the forward pass at the optimal layer (L20): $\mathbf{h}' = \mathbf{h} + \alpha \cdot \hat{\mathbf{w}}$.
The learned correctness direction (green) produces a monotonic 9.1 percentage point swing: $\alpha = -5$ raises the error rate to 0.61, $\alpha = +5$ lowers it to 0.52, from a baseline of 0.56.
Random ($\mathbf{r} \sim \mathcal{N}(0, I)$, gray, $-0.8$pp) and orthogonal ($\mathbf{r}_\perp$, orange, $-0.3$pp) controls remain at baseline.
Five additional models in Table~\ref{tab:steering_multimodel}; six in total with this one.}
\label{fig:steering}
\end{figure*}

\textbf{Steering shows the direction is causally load-bearing, not that the subspace is low-rank.}
We add the L2-normalized probe weight vector at every token position at the optimal layer during generation, $\mathbf{h}' = \mathbf{h} + \alpha\hat{\mathbf{w}}$, following \citet{li2023inferencetime}, and sweep $\alpha \in [-5,5]$ on held-out TruthfulQA questions (Appendix~\ref{app:steering}).
The effect is monotonic and 9.1pp end to end: the error rate rises from 0.56 to 0.61 at $\alpha = -5$ and falls to 0.52 at $\alpha = +5$ (Figure~\ref{fig:steering}).
Controls localize the effect to the learned direction rather than to the perturbation itself: random directions move the error rate by $-0.8$pp and orthogonal directions by $-0.3$pp.
Five additional instruction-tuned models across the Qwen, Llama, and Gemma families replicate the pattern; over all six, learned-direction effects run $+2.8$ to $+10.0$pp with specificity (learned minus orthogonal) of $+4.1$ to $+9.6$pp (Appendix Table~\ref{tab:steering_multimodel}, $n = 617$ each).
This is a claim about \emph{one} direction at \emph{one} layer.
It establishes that the direction our probe recovers participates in producing the output, and it says nothing about how many dimensions the causal mechanism occupies.

\textbf{Erasure shows necessity, and also does not bound rank.}
RLACE \citep{ravfogel2022linear} projects the representation onto the nullspace of the correctness direction.
Across all 5 tested models this drops AUC to chance (GPT-2 0.48, GPT-2-Large 0.51, Qwen2-7B 0.51, Mistral-7B 0.50, Llama-3B 0.48), so the subspace we identify is necessary for detection and not a redundant correlate.
But for a binary linear concept, rank-1 erasure is already theoretically sufficient \citep{ravfogel2022linear}: a method that succeeds by removing one direction cannot distinguish a 1D mechanism from an 8D one.
INLP \citep{ravfogel2020null} degrades progressively, with the first direction accounting for the largest single drop (Appendix~\ref{app:inlp}), which is consistent with a low-rank mechanism but equally consistent with a 1D one.

\begin{table}[t]
\centering
{\small
\setlength{\tabcolsep}{5pt}
\begin{tabular}{@{}lccccc@{}}
\toprule
\textbf{Model} & \textbf{Peak} & \textbf{AUC} & \textbf{PLS Peak} & \textbf{32D} & $\downarrow$\textbf{32D?} \\
\midrule
GPT-2      & 5D & 0.762 & 3D & 0.735 & Yes \\
GPT-2-L    & 1D & 0.784 & 3D & 0.750 & Yes \\
Llama-3B   & 3D & 0.944 & 5D & 0.935 & No \\
Mistral-7B & 3D & 0.927 & 5D & 0.906 & Yes \\
Qwen2-7B   & 3D & 0.916 & 5D & 0.894 & Yes \\
\bottomrule
\end{tabular}
}
\caption{DAS causal dimension sweep (TruthfulQA, AUC). Peaks at 1--5D. $\downarrow$32D = degradation at 32D.}
\label{tab:das_dim}
\end{table}

\textbf{Only DAS bounds the causal rank.}
Distributed Alignment Search \citep{wu2023interpretability, geiger2024causal} learns a rotation $\mathbf{R} \in \mathbb{R}^{d \times k}$ under an interchange-intervention objective: swapping the $k$-dimensional projected representation between a correct and an incorrect example should flip the prediction.
Because $k$ is swept explicitly ($k \in \{1,2,3,4,5,8,12,16,32\}$ at the optimal layer), the sweep is what licenses a statement about rank.
DAS is heavier than PLS, so we run it on 5 models spanning the full scale range (GPT-2 124M to Qwen2-7B; Table~\ref{tab:das_dim}).
Peaks land at 1--5D and 4 of 5 models degrade at 32D, so for these models the causal effect is carried by at most 5 dimensions.
Exact peaks differ from the PLS sweep (GPT-2: DAS 5D vs PLS 3D) because DAS optimizes an intervention loss while PLS maximizes label covariance; the agreement is qualitative, not point-for-point.

The direction is therefore efficacious (steering) and necessary (erasure), and the causal effect fits in $\leq$5 dimensions (DAS). The rank claim rests on DAS alone, bounded by the 5 models and the layer we tested rather than established in general.

\subsection{Internal Representations vs Output-Based Methods}
\label{sec:res_internal}

\begin{table}[t]
\centering
{\small
\setlength{\tabcolsep}{4pt}
\begin{tabular}{@{}lcccccr@{}}
\toprule
\textbf{Model} & \textbf{P(T)} & \textbf{Ent.} & \textbf{SE} & \textbf{CCS} & \textbf{Probe} & $\Delta$ \\
\midrule
GPT-2      & 0.44 & 0.51 & 0.56 & 0.54 & \textbf{0.80} & +48\% \\
GPT-2-Med  & 0.45 & 0.51 & 0.60 & 0.61 & \textbf{0.84} & +38\% \\
Mistral-7B & 0.53 & 0.51 & 0.55 & 0.85 & \textbf{0.92} & +8\% \\
Qwen2-7B   & 0.64 & 0.52 & 0.58 & 0.80 & \textbf{0.94} & +18\% \\
\bottomrule
\end{tabular}
}
\caption{Internal probes vs output methods on TruthfulQA (GroupKFold AUC). Probe standard deviations are given in Table~\ref{tab:main_results}.}
\label{tab:baselines}
\end{table}

The internal signal does not uniformly beat reading the output distribution; the advantage is confined to one regime.
On TruthfulQA, output methods (P(True), token entropy, semantic entropy) fall to 0.44--0.64 while probes reach 0.80--0.94 (Table~\ref{tab:baselines}).
On standard QA the gap closes: across four datasets on Qwen2-7B, P(True) reaches 0.93--0.95 on SciQ, CSQA, and FEVER, comparable to centroid distance at 0.91--0.97, and only TruthfulQA produces a large separation (P(True) 0.48 vs centroid 0.93; Appendix~\ref{app:multi_dataset_baselines}).
The reason is specific to adversarial misconception data \citep{lin2022truthfulqa}: the model confidently asserts the wrong answer, so any signal derived from the output distribution encodes the belief rather than its correctness.
The same regime breaks the unsupervised probe CCS \citep{burns2023discovering}, which reaches 0.54--0.85 against 0.80--0.94 for supervised probes.
The claim we carry forward is therefore centroid sufficiency, not internal superiority.

\textbf{Comparison to a single-pass detector.}
We ran LLM-Check \citep{sriramanan2024llmcheck} on TruthfulQA, applying its Hidden Score (mean log singular value of the centered token-Gram matrix) to GPT-2-Large over an 8-layer sweep ($\ell \in \{5,10,15,20,23,25,30,35\}$, $n = 1556$, sign-flipped).
Its best AUC is 0.53 at $\ell = 15$, with all layers in 0.513--0.528, effectively chance, against 0.79 for centroid distance in the same 8D PLS subspace on the same model and dataset: a 26-point gap.
The two methods answer different questions, and we read the gap as a design difference rather than a defect in either.
LLM-Check scores sequence-level activation abnormality without labels, which is what free-form hallucination detection requires; our centroid uses labeled contrastive pairs at a single token position, which is available in the teacher-forced setting and not in deployment.

\subsection{Transfer Requires Joint Training}
\label{sec:res_transfer}

\begin{table}[t]
\centering
{\small
\setlength{\tabcolsep}{4pt}
\begin{tabular}{@{}lcccc@{}}
\toprule
\textbf{Model} & \textbf{In-Dom} & \textbf{Single} & \textbf{Multi} & $\boldsymbol{\Delta}$ \\
\midrule
GPT-2          & 0.732 & 0.500 & 0.503          & $+0.3$ \\
GPT-2-Med      & 0.750 & 0.501 & 0.538          & $+3.7$ \\
GPT-2-Large    & 0.751 & 0.509 & 0.561          & $+5.2$ \\
Llama-3.2-1B   & 0.924 & 0.621 & \textbf{0.729} & $+10.8$ \\
Qwen2-1.5B     & 0.877 & 0.532 & \textbf{0.848} & $+31.6$ \\
Qwen2-7B       & 0.901 & 0.660 & \textbf{0.905} & $+24.5$ \\
Mistral-7B     & 0.930 & 0.394 & \textbf{0.806} & $+41.2$ \\
Llama-3.2-3B   & 0.970 & 0.648 & \textbf{0.799} & $+15.0$ \\
\bottomrule
\end{tabular}
}
\caption{Cross-domain generalization (AUC, fixed PLS-5D), averaged over SciQ, CSQA and FEVER (HaluEval excluded; see text). \textbf{Single}: probe trained on TruthfulQA only. \textbf{Multi}: leave-one-out multi-dataset training. Per-dataset breakdown in Appendix~\ref{app:cross_dataset_full}.}
\label{tab:cross_dataset}
\label{tab:multi_dataset}
\end{table}

A subspace that is clean in-domain is not thereby a transferable one.
A probe fit on TruthfulQA alone transfers close to random: 0.39--0.66 AUC on SciQ, CSQA, and FEVER, with the GPT-2 family at 0.50--0.51 and Qwen2-7B highest at 0.66 (Table~\ref{tab:cross_dataset}; per-dataset in Appendix~\ref{app:cross_dataset_full}).
The table covers the 8 models for which cross-dataset embeddings were extracted; the other three are absent for that reason, not by selection.

The failure is one of orientation, not of absence.
Refitting PLS-8 natively on each dataset recovers the same low-rank structure in-domain everywhere (TruthfulQA 0.71--0.94, SciQ 0.63--0.96, FEVER 0.79--0.95, CSQA 0.61--0.89 over 8 models; Table~\ref{tab:indomain_4datasets}, Appendix~\ref{app:multi_dataset_indomain}), so each dataset does contain a decodable correctness signal.
HaluEval makes the point sharply: a TruthfulQA-trained probe scores 0.18--0.47 on it, \emph{below} chance, yet refitting natively on HaluEval yields 0.99 in-domain AUC, and the two directions are essentially orthogonal ($|\cos\theta| = 0.08$).
HaluEval's question/passage/answer triples are structurally unlike TruthfulQA's misconception/correct pairs, and a below-chance score is the signature of a direction that is present but pointed elsewhere.
We exclude HaluEval from the cross-domain average for this reason.

Because the problem is a mismatch of direction, the fix is to constrain the subspace with more than one dataset rather than to regularize harder.
Training jointly on three datasets (leave-one-out), following the multi-dataset protocol of \citet{burns2023discovering, marks2024geometry} and the Universal Truthfulness Hyperplane of \citet{liu2024universal}, lifts transfer on all five instruction-tuned models tested: Qwen2-7B $+31$/$+28$pp on SciQ/CSQA, Mistral-7B $+42$/$+33$pp, Llama-3B $+24$/$+23$pp (Table~\ref{tab:multi_dataset}).
The gain is bounded by model class rather than by protocol: the GPT-2 base models gain only $+0.3$ to $+5.2$pp under identical treatment.
We report this boundary as a limit on the recipe, not a claim about capacity: model class and scale are confounded in our grid (\S\ref{sec:discussion}).

\subsection{Where the Signal Sits, and What Could Confound It}
\label{sec:res_arch}

Depth and scale behave regularly.
Within a family, AUC increases with scale (GPT-2 0.80 $\rightarrow$ GPT-2-Large 0.84; Qwen2-7B 0.94 $\rightarrow$ Qwen2.5-14B 0.95), while the optimal layer moves earlier as models grow: the GPT-2 family peaks at its final layer, larger models at 46--75\% depth (Table~\ref{tab:main_results}).
The embedding manifold compresses from 20--55D at early layers to 8--12D at the optimal layer, and cross-model variance falls with depth (std 0.28 early, 0.11 late; Appendix~\ref{app:dimension_evolution}), so five architecture families converge on a similar discriminative dimensionality despite differing early-layer geometry.
Cross-layer probe-weight similarity is block-diagonal (mean intra-phase similarity 0.81; Appendix~\ref{app:geometry}): early layers extract token features, middle layers integrate semantics, and late layers make the correctness signal linearly available.
Dimensionality alone does not explain quality, however: lower dimension correlates with higher AUC at $r = -0.43$ but accounts for only 18\% of variance.
Orientation matters more, which is why PLS beats both LDA, capped at 1D for a binary target, and unsupervised reduction across all 11 models (Appendices~\ref{app:pls_lda},~\ref{app:compression}).
Probes do not transfer between models in any useful absolute sense: within the GPT-2 family, a GPT-2 probe reads GPT-2-Medium activations at 0.59 AUC against a 0.68 native ceiling under the same protocol, the reverse direction collapses to chance (0.50), and cross-family transfer through a common-dimension projection (GPT-2 $\leftrightarrow$ Qwen2-7B) sits at or below chance (0.44--0.48).
What is shared across models is the abstract structure of the encoding, its dimensionality and mean-shift form, not the parameters of any particular probe.

\textbf{Confounds.}
The probe is not reading surface form.
A length-only probe reaches 0.54 AUC ($r = -0.016$, $p = 0.52$) and length balancing costs 1.3\%, while correlations with embedding statistics (L2 norm, mean activation, sparsity) stay below $|r| = 0.15$.
Answer-string plausibility is ruled out by a paraphrase control: 5 paraphrase variants per answer across 817 TruthfulQA questions cluster rather than scatter (between/within-answer $F = 17.40$, $p < 0.001$), and GroupKFold test AUC on unseen paraphrases is 0.926 (Appendix~\ref{app:paraphrase}).

\textbf{Teacher-forced scope.}
All results above are teacher-forced, standard in this line of work \citep{marks2024geometry, burns2023discovering, orgad2025llmsknow}, and the right setting for asking whether correctness is \emph{encoded}.
It is not the deployment setting.
Extending them to a model's own free-form generations requires an automatic correctness judge, and judge choice is known to shift measured detection performance substantially \citep{santilli2025revisiting}; we treat that extension as open (\S\ref{sec:discussion}).

\section{Discussion}
\label{sec:discussion}

\textbf{A low-rank reading of the linear representation hypothesis.}
The discriminative signal occupies 2--8 dimensions, well below the 8--12D intrinsic dimension of the embedding manifold it sits inside.
The linear representation hypothesis \citep{park2024the} states that features are directions; correctness turns out to be more constrained.
It is separable by a hyperplane between two class means inside a subspace an order of magnitude smaller than the manifold: a low-rank linear separator, not a direction in the full residual stream (Appendix~\ref{app:instruct_small_manifold} visualizes the 3D slice).
The block-diagonal layer structure (\S\ref{sec:res_arch}) gives this a phase-specific reading: the signal becomes linearly available only in the late phase.

\textbf{What is new here, and what is not.}
Two of our load-bearing observations are established, and we do not claim them.
\citet{marks2024geometry} report that difference-of-means probing matches or exceeds trained linear probes for truth directions on contrastive pairs.
\citet{burger2024truth} likewise establish low-dimensional linear separability, identifying a small universal truth subspace shared across several 7B--13B models.
Read against that work, low-rank separability of correctness strengthens a known result rather than establishing a new phenomenon.

We add three things prior work leaves open.
First, \emph{why} two means suffice: separation is a mean shift and not a covariance difference, which is why Mahalanobis distance, kernel SVM, and nearest-convex-hull classifiers all converge to the same AUC \citep{ng2001discriminative}.
The shift is already present in the raw residual stream (PLS adds only $+0.057$), and multi-prototype variants \emph{hurt}: the per-class distributions are unimodal, so two means is the correct model, not merely a sufficient one.
Second, a \emph{bounded causal rank}: prior work shows truth directions exist and can be steered; how many dimensions the mechanism occupies is a separate question, which DAS answers by sweeping $k$ explicitly (\S\ref{sec:res_causal}).
This bound rests on DAS alone: steering and erasure establish efficacy and necessity but are rank-agnostic, so treating all three as joint evidence for low rank would overstate what they show.
Third, \emph{recipe-dependent transfer}: single-dataset directions are near-orthogonal across domains ($|\cos\theta| = 0.08$ between TruthfulQA and HaluEval), so joint training is a structural correction for directional mismatch rather than generic regularization.
Multi-dataset truthfulness training is prior art \citep{liu2024universal}; we contribute the diagnosis of why it is necessary.

\textbf{Relation to unsupervised low-rank detection.}
HaloScope \citep{du2024haloscope} pursues a similar low-rank intuition without labels; our transfer results explain its difficulty: a subspace recovered on one distribution points elsewhere on another, so unsupervised recovery must solve the orientation problem that joint training solves with labels; a head-to-head comparison is left to future work.

\section{Limitations}
\label{sec:limitations}

(i) Models span 124M--14B autoregressive decoders; we make no claim about encoder--decoder or non-transformer architectures.
(ii) Probes are trained and evaluated teacher-forced, standard in this line of work \citep{marks2024geometry, burns2023discovering, orgad2025llmsknow}; the geometry is thus a property of the teacher-forced representation, and extending detection to free-form generations is open, complicated by unreliable correctness judging \citep{santilli2025revisiting}.
(iii) We localize the subspace but not what its dimensions encode; a covariance-maximizing basis is fixed only up to rotation, so component-level semantics needs a decomposition with an identifiability guarantee (\S\ref{sec:res_dim}).
(iv) Causal validation is in-distribution, on each dataset's training partition; cross-distribution causal transfer is untested.
(v) The transfer recipe's model-class boundary is confounded: every model we test below 1B is a base model and every model at 1B or above is instruction-tuned, so scale and instruction tuning cannot be separated in our grid.
(vi) Correctness is treated as binary; compositionality and negation are unaddressed, and negation is where \citet{burger2024truth} report additional structure.
(vii) Steering is dual-use: the direction that enables detection also enables manipulation.

\section{Conclusion}
\label{sec:conclusion}

Across eleven models from five families, correctness occupies a 2--8D subspace of a residual stream at least 768 dimensions wide, and detection collapses to two class means. Steering and erasure show the direction is causal and necessary; only DAS bounds its rank. Single-dataset probes need joint training to transfer; the dimensions' semantics and the free-form generation gap remain open.

\bibliography{references}
\bibliographystyle{icml2026}

\onecolumn
\appendix
\appendix

\section{Reproducibility and Implementation}
\label{app:reproducibility}

\subsection{Seed Robustness}
\label{app:seeds}

Classification experiments use multiple random seeds (42, 123, 456) controlling stratified train/test splits, GroupKFold assignment by question ID, and stochastic generation for semantic entropy. Causal experiments (steering, RLACE, DAS) fix seed=42 and rely on the large held-out sample (n=617 for steering) for statistical power instead of seed averaging.
We report mean $\pm$ std where variance is non-trivial.

\textbf{Variance analysis.} Most metrics show low variance: probe AUC std $< 0.02$ for 10/11 models, in-domain transfer std $< 0.01$.
Higher variance appears in Mistral-7B/Gemma-2B cross-dataset transfer (std $\approx 0.20$), potentially reflecting dataset-model mismatch.
All key comparisons (probe vs.\ entropy) achieve $p < 0.05$ after Bonferroni correction via paired t-tests across seeds.

\subsection{Implementation Details}
\label{app:implementation}

\textbf{Models.} GPT-2 (124M, 12L), GPT-2-Medium (355M, 24L), GPT-2-Large (774M, 36L), Qwen2-1.5B-Instruct (28L), Qwen2-7B-Instruct (28L), Llama-3.2-1B-Instruct (16L), Llama-3.2-3B-Instruct (28L), Llama-3.1-8B-Instruct (32L), Qwen2.5-14B-Instruct (48L), Mistral-7B-Instruct-v0.3 (32L), Gemma-2-2B-it (26L).

\textbf{Extraction.} Residual stream activations at last-token position following \citet{gurnee2023finding}.
PLS dimension reduction treats labels as continuous targets (0/1).

\textbf{Probing.} Logistic regression ($C=0.1$); 640--1,280 samples per dataset; 5-fold GroupKFold CV grouped by question ID to prevent data leakage.
Layer and PLS dimension selection use CV test AUC (averaged across held-out folds), not train AUC, ensuring no information leakage from hyperparameter selection.

\textbf{SAE analysis.} SAELens with Neuronpedia pretrained GPT-2 models (24,576 features, 32$\times$ expansion).

\section{Geometric Analysis of the Discriminative Subspace}
\label{app:geometry}

This section provides detailed geometric characterization of how confidence representations evolve through transformer layers.
Our analysis reveals three key findings: (1) intrinsic dimension follows a compression pattern: initially expanding in early layers (peaking around 10--20\% depth) before decreasing through middle and late layers, (2) probe weight similarity exhibits block-diagonal structure indicating distinct processing phases, and (3) dimension alone explains only 18\% of classification variance: the \emph{orientation} of the low-dimensional subspace matters more than its dimensionality.

\subsection{Compression Pattern}
\label{app:compression}

\begin{figure*}[t]
\centering
\includegraphics[width=\textwidth]{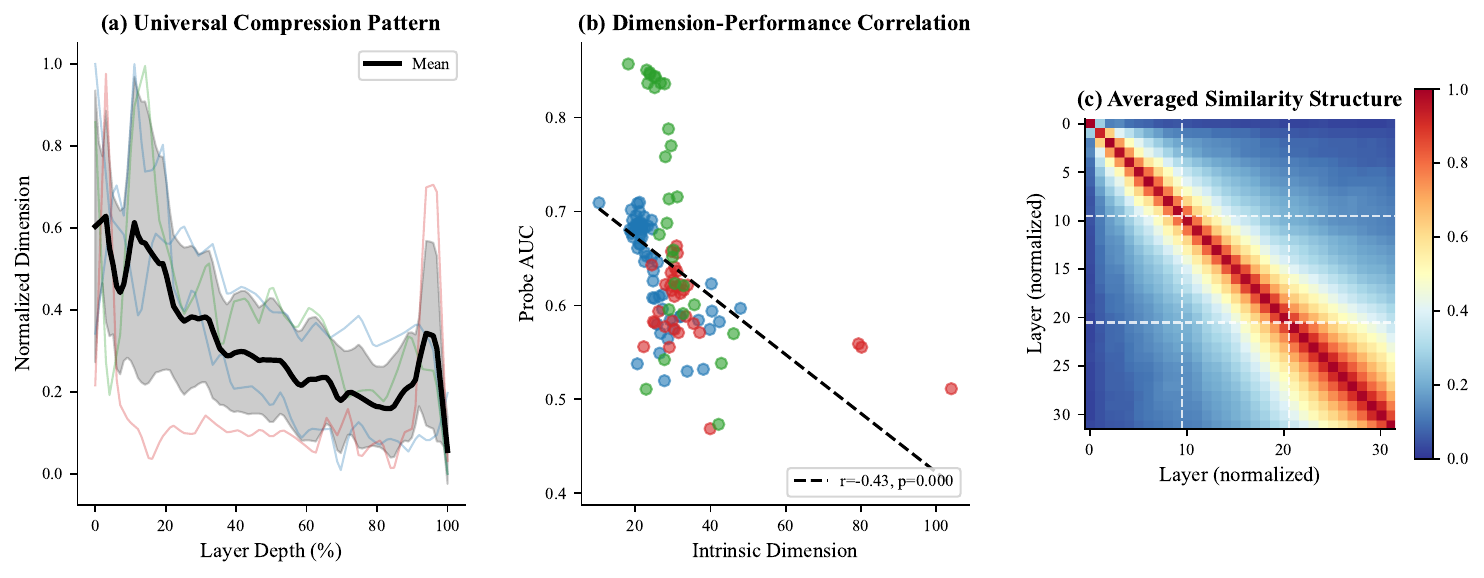}
\caption{\textbf{Geometric patterns across architectures.} (a)~Normalized intrinsic dimension (MLE) by layer depth.
All models compress from early to late layers (mean curve in black), with peak dimension at 10--20\% depth. (b)~Dimension-performance correlation: lower intrinsic dimension correlates with higher probe AUC ($r = -0.43$, $p < 0.001$), but $R^2 = 0.18$ indicates dimension explains less than one-fifth of variance; classification utility depends on direction, not dimensionality. (c)~Cross-layer probe weight similarity averaged across models shows three-phase block-diagonal structure with phase boundaries at 30\% and 70\% depth.}
\label{fig:universal_geometry}
\end{figure*}

Figure~\ref{fig:universal_geometry} synthesizes geometric properties across all models, revealing consistent patterns in how confidence is encoded.

\textbf{Compression dynamics.} Intrinsic dimension (estimated via Levina-Bickel MLE \citep{levina2004maximum}) decreases monotonically from 20--55D at early layers to 8--12D at optimal layers, a 40--60\% reduction.
This compression is consistent across families: GPT-2, Mistral, Qwen, Llama, and Gemma families all converge to similar final dimensionality despite vastly different training procedures and scales.
Early layers (0--20\% depth) show high cross-model variance (std = 0.28 normalized units); late layers converge (std = 0.11), suggesting that while models initialize representations differently, they converge to similar compressed confidence encodings.

\textbf{Dimension vs.\ performance.} The negative correlation ($r = -0.43$) between intrinsic dimension and probe AUC initially suggests that lower-dimensional representations yield better classification.
However, the weak $R^2 = 0.18$ reveals that dimension is \emph{necessary but not sufficient}: the \emph{orientation} of the low-dimensional subspace relative to the correct/incorrect decision boundary matters more than its raw dimensionality.
A 10D subspace aligned with the confidence direction outperforms a 5D subspace misaligned with it.
This explains why PLS outperforms unsupervised dimension reduction: PLS finds the 2--8D subspace that maximizes class separation, not merely the directions of highest variance. The hunchback compression profile itself mirrors the intrinsic-dimension dynamics \citet{ansuini2019intrinsic} report for deep vision networks; what is specific to our setting is where the discriminative subspace sits inside it.

\textbf{Three-phase processing.} The averaged cross-layer similarity matrix (Figure~\ref{fig:universal_geometry}c) reveals block-diagonal structure with consistent phase boundaries across architectures:
\begin{itemize}[itemsep=1pt,topsep=2pt,leftmargin=*]
    \item \textbf{Phase I (0--30\% depth)}: Token-level feature extraction; low similarity to later layers (mean cross-phase similarity: 0.18)
    \item \textbf{Phase II (30--70\% depth)}: Semantic integration; gradual probe weight rotation (mean within-phase similarity: 0.58)
    \item \textbf{Phase III (70--100\% depth)}: Stable confidence encoding; high intra-phase coherence (mean within-phase similarity: 0.81)
\end{itemize}

\subsection{Architecture-Specific Dimension Evolution}
\label{app:dimension_evolution}

While the compression pattern is consistent, architecture-specific variations provide insights into how different models encode confidence.

\textbf{Consistent compression, model-specific timing.} The layer sweep tracks intrinsic dimension through network depth for four representative models.
Smaller models (124M--774M) show earlier phase transitions (boundaries at 25\%/60\%) than the 1B--7B models (30\%/70\%), suggesting larger models prolong rich intermediate representations before final compression.

\textbf{Mistral anomaly.} Mistral-7B exhibits late-layer dimension \emph{expansion} from 25D at L25 to 80+D at L30--32.
This expansion correlates with unembedding preparation: Mistral's architecture appears to re-expand representations before projecting to vocabulary space.
This expansion does \emph{not} improve classification; Mistral's optimal layer is L23 (72\% depth), \emph{before} the expansion begins.
This confirms that the confidence signal crystallizes in middle layers; late-layer expansion serves output generation, not confidence encoding.

\subsection{Confidence Landscape}
\label{app:3d_surface}

To understand the interaction between layer selection and dimensionality, we sweep the full layer-by-dimension grid for our best-performing model.

\textbf{Ridge structure.} A clear AUC ridge runs through the surface at 2--8 PLS dimensions across all layers.
Performance degrades sharply below 2D (insufficient capacity to capture the signal) and above 16D (overfitting to training noise).
The ridge is narrower at early layers (optimal: 3--4D) and broader at late layers (optimal: 4--8D), reflecting increased signal-to-noise ratio in later representations.

\textbf{Layer dominates dimension.} Quantifying the relative importance: fixing dimension at 5D, AUC varies from 0.52 (L0) to 0.92 (L23), a 77\% relative improvement.
Fixing layer at L23, AUC varies from 0.86 (1D) to 0.92 (5D), only 7\% improvement.
\textbf{Layer selection is 11$\times$ more impactful than dimension selection.} This motivates our recommendation to tune layer first, then dimension, rather than joint optimization.

\section{Geometric Classification Methods}
\label{app:geometric_methods}

Given the low-dimensional structure revealed in Section~\ref{app:geometry}, we investigate whether geometric classifiers can exploit non-linear patterns within the PLS subspace.
Our negative result (geometric methods do not outperform linear probes) provides evidence that within this dominant discriminative subspace, the confidence signal is \emph{linearly separable}.

\subsection{Method Comparison}
\label{app:geometric_comparison}

We evaluate five geometric approaches in 8D PLS space:

\textbf{Linear probe (0.773 AUC).} Standard logistic regression on PLS-reduced activations.
Serves as the discriminative baseline.

\textbf{Centroid distance (0.771 AUC).} Generative approach: classify based on distance to class centroids.
Performance nearly matches the discriminative probe, confirming that class means capture the discriminative signal.
The decision boundary is perpendicular to the line connecting centroids, geometrically equivalent to the probe's learned hyperplane.

\textbf{Local density (0.701 AUC).} Estimate confidence via kernel density ratio $p(\text{correct}|x) / p(\text{incorrect}|x)$ using Gaussian KDE with Scott's rule bandwidth selection.
Underperforms because correct and incorrect distributions have similar local densities: they differ in \emph{location} (centroid position), not \emph{shape} (density profile).

\textbf{KNN-10 (0.748 AUC).} Classify by majority vote of 10 nearest neighbors.
Despite local adaptivity, underperforms linear methods, suggesting the discriminative subspace is globally linear rather than exhibiting local curvature.

\textbf{Ensemble (0.764 AUC).} Average predictions across methods.
No improvement over the best single method, indicating the errors are correlated rather than complementary.

\textbf{Implications.} The equivalence of discriminative (probe) and generative (centroid) approaches reveals that confidence is encoded as a simple \emph{mean shift} in activation space, not a complex decision boundary.
This supports interpretability: a single direction suffices to extract the confidence signal.

\subsection{Cross-Model Geometric Classifier Comparison}
\label{app:geometric_classifiers_table}

Table~\ref{tab:geometry_classifiers_app} reports geometric classifier AUC across 9 models in the 8D PLS subspace.
All methods (linear probe, centroid, Mahalanobis, nearest convex hull, KNN, kernel SVM) converge within $\pm$0.02 AUC, confirming the mean-shift structure across architectures.

\begin{table*}[h]
\centering
\small
\caption{Geometric classifiers in 8D PLS space (GroupKFold AUC).
NCH = Nearest Convex Hull.
Llama-8B and Qwen2.5-14B added after geometry experiments.}
\label{tab:geometry_classifiers_app}
\begin{tabular}{@{}lcccccc@{}}
\toprule
\textbf{Model} & \textbf{Linear} & \textbf{Centroid} & \textbf{Mahal.} & \textbf{NCH} & \textbf{KNN} & \textbf{SVM} \\
\midrule
GPT-2 & 0.757$_{\pm.038}$ & \textbf{0.766}$_{\pm.029}$ & 0.762$_{\pm.036}$ & 0.714$_{\pm.045}$ & 0.752$_{\pm.029}$ & 0.743$_{\pm.037}$ \\
GPT-2-Med & 0.787$_{\pm.023}$ & \textbf{0.799}$_{\pm.020}$ & 0.796$_{\pm.020}$ & 0.746$_{\pm.033}$ & 0.787$_{\pm.023}$ & 0.779$_{\pm.020}$ \\
GPT-2-Large & 0.777$_{\pm.017}$ & \textbf{0.797}$_{\pm.016}$ & 0.791$_{\pm.017}$ & 0.754$_{\pm.028}$ & 0.779$_{\pm.017}$ & 0.773$_{\pm.015}$ \\
Llama-1B & 0.881$_{\pm.016}$ & \textbf{0.885}$_{\pm.012}$ & 0.880$_{\pm.015}$ & 0.849$_{\pm.014}$ & 0.864$_{\pm.018}$ & 0.857$_{\pm.025}$ \\
Qwen2-1.5B & \textbf{0.859}$_{\pm.045}$ & \textbf{0.859}$_{\pm.029}$ & 0.858$_{\pm.036}$ & 0.808$_{\pm.038}$ & 0.847$_{\pm.031}$ & 0.842$_{\pm.036}$ \\
Gemma-2B & 0.846$_{\pm.021}$ & \textbf{0.851}$_{\pm.022}$ & 0.846$_{\pm.032}$ & 0.818$_{\pm.027}$ & 0.817$_{\pm.024}$ & 0.822$_{\pm.026}$ \\
Llama-3B & \textbf{0.922}$_{\pm.019}$ & 0.919$_{\pm.017}$ & 0.918$_{\pm.023}$ & 0.900$_{\pm.019}$ & 0.903$_{\pm.022}$ & 0.906$_{\pm.029}$ \\
Qwen2-7B & 0.885$_{\pm.019}$ & \textbf{0.891}$_{\pm.017}$ & 0.887$_{\pm.020}$ & 0.849$_{\pm.014}$ & 0.875$_{\pm.012}$ & 0.878$_{\pm.015}$ \\
Mistral-7B & 0.890$_{\pm.023}$ & \textbf{0.897}$_{\pm.020}$ & 0.893$_{\pm.022}$ & 0.851$_{\pm.029}$ & 0.883$_{\pm.021}$ & 0.880$_{\pm.028}$ \\
\bottomrule
\end{tabular}
\end{table*}

\subsection{Unsupervised Features}
\label{app:unsupervised}

\begin{table}[h]
\centering
\footnotesize
\caption{Unsupervised features (AUC, no labels).
All fall well short of supervised probes.}
\label{tab:unsupervised}
\begin{tabular}{@{}lccc@{}}
\toprule
\textbf{Feature} & \textbf{GPT-2} & \textbf{Mistral-7B} & \textbf{Llama-1B} \\
\midrule
L2 norm & 0.508$_{\pm0.003}$ & \textbf{0.621}$_{\pm0.008}$ & \textbf{0.608}$_{\pm0.026}$ \\
Recon. error & 0.531$_{\pm0.025}$ & 0.556$_{\pm0.008}$ & 0.546$_{\pm0.010}$ \\
LOF score & 0.520$_{\pm0.017}$ & \textbf{0.572}$_{\pm0.006}$ & 0.556$_{\pm0.034}$ \\
Cluster unc. & 0.523$_{\pm0.016}$ & 0.534$_{\pm0.014}$ & 0.557$_{\pm0.029}$ \\
Local dim & \textbf{0.547}$_{\pm0.038}$ & \textbf{0.578}$_{\pm0.045}$ & \textbf{0.590}$_{\pm0.055}$ \\
\cmidrule{1-4}
Best superv. & 0.757$_{\pm0.038}$ & 0.890$_{\pm0.023}$ & 0.881$_{\pm0.016}$ \\
\bottomrule
\end{tabular}
\end{table}

\section{Baseline Method Details}
\label{app:baselines}

\subsection{Output-Based Baselines}
\label{app:output_baselines}

\textbf{P(True).} Probability the model assigns to ``Yes'' when explicitly asked whether its previous answer is correct.
\textbf{NLL.} Negative log-likelihood of the answer tokens under the model.
\textbf{Token entropy.} $H(p) = -\sum_i p_i \log p_i$ over the next-token distribution.
\textbf{Verbalized confidence.} The model's self-reported confidence on a 1--10 scale, prompted after answer generation.
\textbf{Semantic entropy} \citep{farquhar2024semantic}: entropy over semantically clustered generations ($K{=}5$ samples, NLI clustering; protocol below).

\subsection{Unsupervised Baselines}
\label{app:unsup_baselines}

\textbf{CCS} \citep{burns2023discovering}: contrastive consistency search for truth directions, fit on contrast pairs without correctness labels.
\textbf{L2 norm}: activation magnitude $\|\mathbf{h}\|_2$.
\textbf{Reconstruction error}: residual from a small autoencoder trained on hidden states.
\textbf{LOF score}: local outlier factor for anomaly detection.
\textbf{Cluster uncertainty}: distance to the nearest k-means cluster centroid in activation space.

\subsection{Why Direct Labels, Not Entropy?}
\label{app:why_labels}

A methodological note: probes trained on semantic entropy targets \emph{failed} to separate incorrect from correct samples, while probes trained on direct correctness labels succeed.
The reason is that uncertainty $\neq$ incorrectness.
Models can be confidently wrong (low entropy, hallucinating) or appropriately uncertain (high entropy on ambiguous inputs).
On TruthfulQA, which targets common misconceptions models inherit and assert with high confidence, the entropy/correctness alignment is especially weak (\S\ref{app:se}).
Direct labels capture what we want to detect; entropy proxies do not.

\subsection{Semantic Entropy Analysis}
\label{app:se}

We implement semantic entropy following \citet{farquhar2024semantic} to understand why uncertainty-based methods underperform.

\textbf{Protocol.} (1)~Generate $K=5$ completions per prompt (nucleus sampling, $p=0.9$, $T=0.7$). (2)~Cluster by semantic equivalence via bidirectional NLI (DeBERTa-v3-large): two responses are equivalent if NLI predicts entailment in both directions. (3)~Compute entropy: $\text{SE} = -\sum_c p_c \log p_c$ over cluster probabilities.

\textbf{Results on TruthfulQA.} Mean SE for correct answers: $0.060 \pm 0.089$; for incorrect answers: $0.115 \pm 0.142$.
Cohen's $d = 0.47$ (small-medium effect).
Classification AUC = 0.58, far below probe AUC (0.77--0.92).

\textbf{Why SE underperforms.} TruthfulQA tests \emph{common misconceptions}: questions where humans frequently give wrong answers.
Models inherit these misconceptions and assert them \emph{confidently}.
The 0.055 SE gap confirms incorrect answers are slightly more uncertain on average, but the effect is too weak for reliable detection.
\textbf{Semantic entropy detects \emph{uncertainty}, not \emph{incorrectness}}: these are distinct signals, and TruthfulQA specifically targets confident errors.

\subsection{SAE Feature Analysis}
\label{app:sae}

We analyze whether pretrained Sparse Autoencoders (SAEs) can provide interpretable confidence features.
If confidence localizes to specific SAE features, this would enable mechanistic interpretation of confidence encoding.

\textbf{Setup.} Neuronpedia gpt2-small-res-jb SAE (layer 6, 24,576 features, 32$\times$ expansion).
We use layer 6 (mid-network) because it is the deepest layer for which a pre-trained Neuronpedia SAE is available; the optimal probe layer for GPT-2 is L11 (Table~\ref{tab:main_results}).

\textbf{Feature statistics.} By activation frequency: sparse ($<$1\%): 13 features; moderate (1--10\%): 21,857 features (88.9\%); dense ($>$10\%): 2,706 features (11.0\%).
The heavy tail toward moderate activation suggests most features are contextually specific.

\textbf{Correctness correlation.} Of 24,576 features, only 307 show significant correlation with correctness labels ($p < 0.05$, Bonferroni-corrected $p < 2\times10^{-6}$).
Maximum $|r| = 0.238$.
Top positive features (incorrect-associated): uncertainty markers, hedging language, abstract concepts.
Top negative features (correct-associated): named entities, numerical expressions, specific facts.

\textbf{Limitation.} Individual SAE features explain $<$6\% of variance ($r^2 < 0.057$), while linear probes explain 35\%+.
\textbf{Confidence is distributed across many features}, not localized to interpretable atoms.
This motivates our probe-based approach over feature-based interpretability.

\section{Extended Results}
\label{app:extended}

\subsection{Small Instruction-Tuned Models}
\label{app:instruct_small_manifold}

\begin{figure*}[t]
\centering
\includegraphics[width=0.9\textwidth]{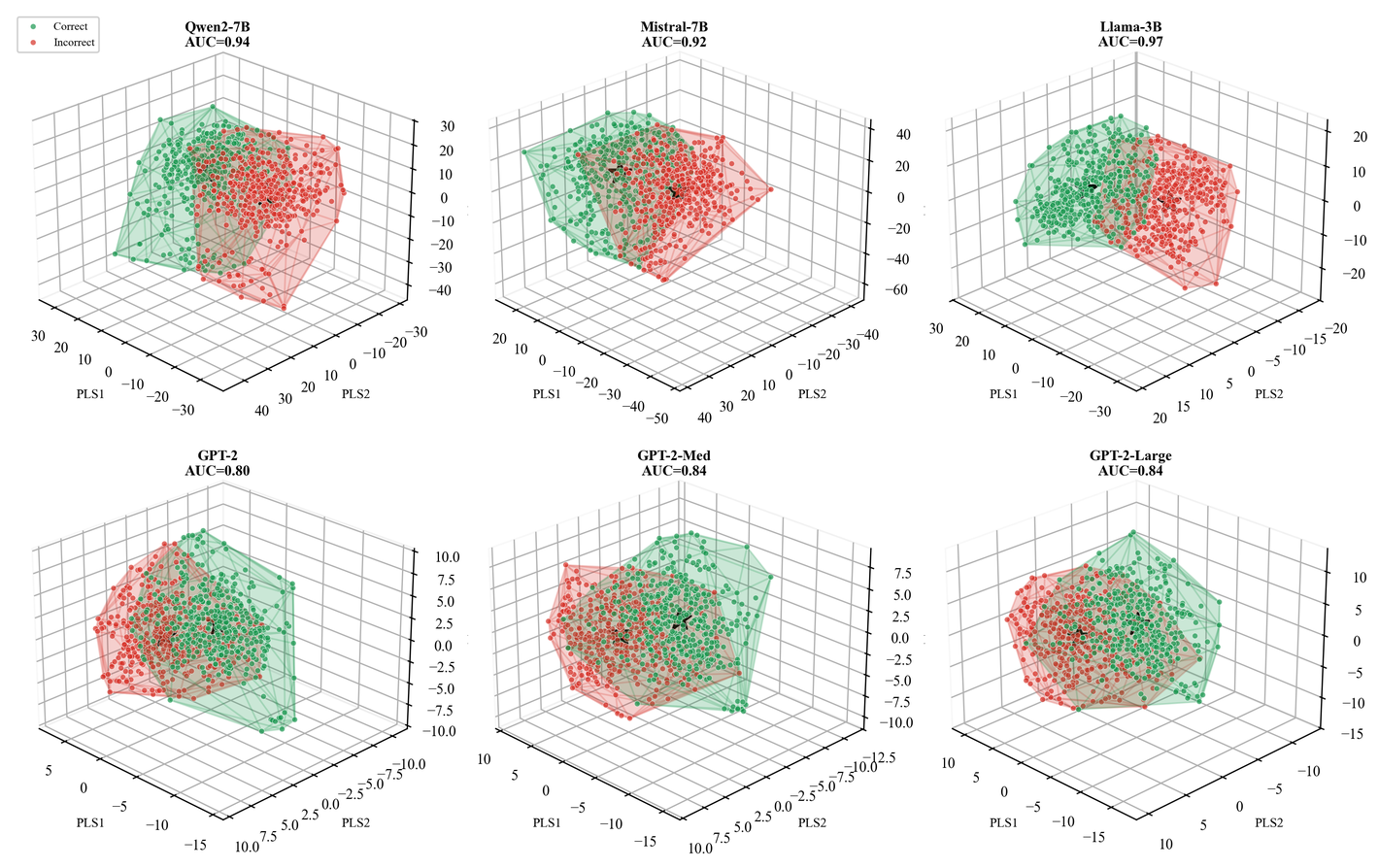}
\caption{3D PLS visualization of the discriminative subspace.
Row 1: instruction-tuned models (Qwen2-7B, Mistral-7B, Llama-3B).
Row 2: GPT-2 family (base models).
Convex hulls show class regions; stars mark centroids.
The GPT-2 family shows clearer visual separation despite lower AUC (0.80--0.84), while instruction-tuned models achieve higher AUC (0.92--0.97) with more overlap in the 3D projection.
Smaller instruction-tuned models are in Appendix~\ref{app:instruct_small_manifold}.}
\label{fig:manifold_3d}
\end{figure*}

\begin{figure}[h]
\centering
\includegraphics[width=0.85\linewidth]{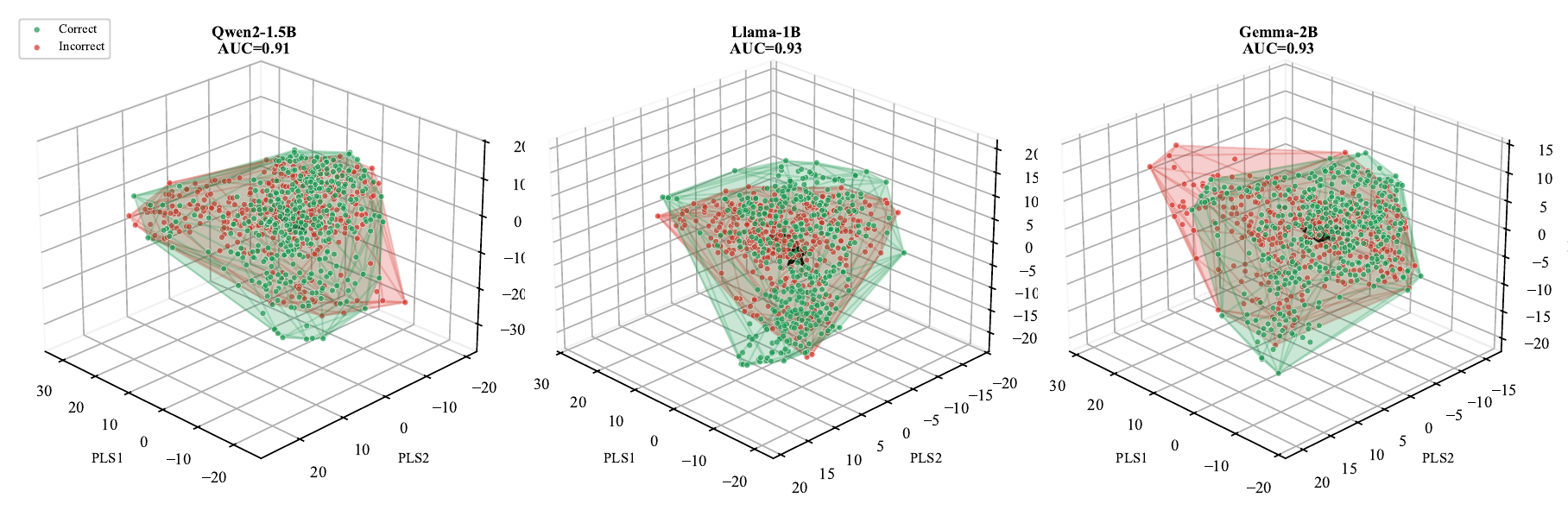}
\caption{3D PLS visualization of the discriminative subspace for the 1B--2B models (Qwen2-1.5B, Llama-1B, Gemma-2B).
These models achieve AUC 0.91--0.93, comparable to the 3B--7B models reported elsewhere in the appendix.
Visual separation appears less distinct due to higher overlap in the projected 3D space, despite strong full-dimensional classification performance.}
\label{fig:instruct_small_manifold}
\end{figure}

\subsection{Layer-wise Performance Table}
\label{app:layerwise}

\begin{table}[h]
\centering
\small
\caption{\textbf{Layer-wise performance across GPT-2 family.} AUC and intrinsic dimension (Dim) at key depth percentiles.
All models show monotonic AUC increase concurrent with dimension compression, supporting the hypothesis that compression and confidence encoding co-occur.}
\label{tab:gpt2_layers}
\begin{tabular}{@{}l rr rr rr@{}}
\toprule
& \multicolumn{2}{c}{\textbf{GPT-2}} & \multicolumn{2}{c}{\textbf{GPT-2-Medium}} & \multicolumn{2}{c}{\textbf{GPT-2-Large}} \\
\cmidrule(lr){2-3} \cmidrule(lr){4-5} \cmidrule(l){6-7}
\textbf{Depth} & AUC & Dim & AUC & Dim & AUC & Dim \\
\midrule
0\% & 0.632 & 24.3 & 0.614 & 32.1 & 0.598 & 41.2 \\
25\% & 0.648 & 18.7 & 0.652 & 24.6 & 0.661 & 28.9 \\
50\% & 0.704 & 14.2 & 0.691 & 18.3 & 0.722 & 21.4 \\
75\% & 0.670 & 11.5 & 0.738 & 12.1 & 0.768 & 14.7 \\
\rowcolor{bestrow}
100\% & \textbf{0.772} & \textbf{8.9} & \textbf{0.799} & \textbf{8.3} & \textbf{0.812} & \textbf{11.8} \\
\bottomrule
\end{tabular}
\end{table}

\subsection{Dimension Sweep with Fold Variability}
\label{app:dimension_sweep_std}

Table~\ref{tab:dimension_sweep_std} replicates the main-text dimension sweep with per-fold standard deviations (5-fold GroupKFold). Standard deviations grow with dimension on most models (e.g.\ GPT-2: $\pm.003$ at 1D vs $\pm.034$ at 32D), consistent with the interpretation that dimensions beyond the peak add estimation variance rather than signal.

\begin{table}[h]
\centering
\small
\setlength{\tabcolsep}{3.5pt}
\caption{\textbf{PLS dimension sweep with standard deviations} (GroupKFold AUC, TruthfulQA).}
\label{tab:dimension_sweep_std}
\begin{tabular}{@{}lrrrrrrrr@{}}
\toprule
\textbf{Model} & \textbf{1D} & \textbf{2D} & \textbf{3D} & \textbf{4D} & \textbf{5D} & \textbf{8D} & \textbf{16D} & \textbf{32D} \\
\midrule
GPT-2          & 0.718$_{\pm.003}$ & 0.750$_{\pm.019}$ & 0.758$_{\pm.016}$ & 0.753$_{\pm.019}$ & 0.746$_{\pm.023}$ & 0.720$_{\pm.028}$ & 0.672$_{\pm.048}$ & 0.621$_{\pm.034}$ \\
GPT-2-Med      & 0.747$_{\pm.009}$ & 0.781$_{\pm.011}$ & 0.790$_{\pm.011}$ & 0.784$_{\pm.012}$ & 0.778$_{\pm.014}$ & 0.763$_{\pm.025}$ & 0.723$_{\pm.031}$ & 0.676$_{\pm.036}$ \\
GPT-2-Large    & 0.750$_{\pm.010}$ & 0.784$_{\pm.015}$ & 0.791$_{\pm.011}$ & 0.783$_{\pm.016}$ & 0.769$_{\pm.013}$ & 0.741$_{\pm.018}$ & 0.700$_{\pm.022}$ & 0.673$_{\pm.020}$ \\
Llama-3.2-1B   & 0.847$_{\pm.015}$ & 0.886$_{\pm.015}$ & 0.892$_{\pm.013}$ & 0.895$_{\pm.017}$ & 0.893$_{\pm.015}$ & 0.878$_{\pm.025}$ & 0.831$_{\pm.038}$ & 0.813$_{\pm.041}$ \\
Qwen2-1.5B     & 0.814$_{\pm.021}$ & 0.848$_{\pm.017}$ & 0.858$_{\pm.018}$ & 0.863$_{\pm.020}$ & 0.864$_{\pm.026}$ & 0.873$_{\pm.031}$ & 0.842$_{\pm.033}$ & 0.810$_{\pm.041}$ \\
Gemma-2-2B     & 0.817$_{\pm.030}$ & 0.846$_{\pm.024}$ & 0.861$_{\pm.027}$ & 0.860$_{\pm.029}$ & 0.856$_{\pm.036}$ & 0.830$_{\pm.047}$ & 0.780$_{\pm.052}$ & 0.751$_{\pm.053}$ \\
Llama-3.2-3B   & 0.887$_{\pm.018}$ & 0.914$_{\pm.016}$ & 0.917$_{\pm.023}$ & 0.917$_{\pm.020}$ & 0.919$_{\pm.028}$ & 0.910$_{\pm.032}$ & 0.893$_{\pm.039}$ & 0.881$_{\pm.038}$ \\
Qwen2-7B       & 0.828$_{\pm.011}$ & 0.882$_{\pm.009}$ & 0.892$_{\pm.010}$ & 0.894$_{\pm.011}$ & 0.896$_{\pm.011}$ & 0.890$_{\pm.015}$ & 0.876$_{\pm.021}$ & 0.843$_{\pm.030}$ \\
Mistral-7B     & 0.860$_{\pm.019}$ & 0.888$_{\pm.014}$ & 0.899$_{\pm.014}$ & 0.902$_{\pm.016}$ & 0.902$_{\pm.017}$ & 0.899$_{\pm.019}$ & 0.876$_{\pm.023}$ & 0.849$_{\pm.030}$ \\
Llama-3.1-8B   & 0.895$_{\pm.022}$ & 0.931$_{\pm.024}$ & 0.929$_{\pm.023}$ & 0.929$_{\pm.025}$ & 0.923$_{\pm.028}$ & 0.910$_{\pm.037}$ & 0.900$_{\pm.032}$ & 0.890$_{\pm.033}$ \\
Qwen2.5-14B    & 0.927$_{\pm.015}$ & 0.951$_{\pm.016}$ & 0.951$_{\pm.019}$ & 0.951$_{\pm.018}$ & 0.949$_{\pm.017}$ & 0.943$_{\pm.021}$ & 0.938$_{\pm.023}$ & 0.939$_{\pm.023}$ \\
\bottomrule
\end{tabular}
\end{table}

\subsection{Full Cross-Dataset Results}
\label{app:cross_dataset_full}

Table~\ref{tab:cross_dataset_full} provides complete cross-dataset transfer results across all models and datasets.
Key observations:

\begin{itemize}[itemsep=1pt,topsep=2pt,leftmargin=*]
    \item \textbf{In-domain performance}: All 1B+ models achieve $>$0.87 AUC on TruthfulQA; GPT-2 family achieves 0.73--0.78 AUC
    \item \textbf{HaluEval anomaly}: Cross-domain transfer to HaluEval is \emph{below random} for most models (0.18--0.33 AUC); GPT-2-Large is near-random (0.47), suggesting HaluEval's task structure differs fundamentally from factuality assessment for instruction-tuned models
    \item \textbf{FEVER transfer}: Best cross-domain performance (0.51--0.76 AUC), likely because FEVER's fact verification task is closest to TruthfulQA's factuality assessment
\end{itemize}

\begin{table}[h]
\centering
\small
\caption{\textbf{Complete cross-dataset transfer results.} \emph{In-domain}: train and test on the same dataset (5-fold GroupKFold).
\emph{Cross-domain}: train on TruthfulQA, test on others (zero-shot transfer).
Values are AUC $\pm$ std across seeds.}
\label{tab:cross_dataset_full}
\begin{tabular}{@{}lcccccc@{}}
\toprule
\textbf{Model} & \textbf{TruthfulQA} & \textbf{SciQ} & \textbf{CSQA} & \textbf{HaluEval} & \textbf{FEVER} \\
\midrule
\multicolumn{6}{l}{\textit{In-domain evaluation}} \\
Qwen2-7B & 0.901$_{\pm0.002}$ & 0.953$_{\pm0.004}$ & 0.897$_{\pm0.016}$ & 0.984$_{\pm0.0001}$ & 0.934$_{\pm0.001}$ \\
Qwen2-1.5B & 0.877$_{\pm0.001}$ & 0.883$_{\pm0.014}$ & 0.819$_{\pm0.0002}$ & 0.986$_{\pm0.00002}$ & 0.898$_{\pm0.001}$ \\
Llama-1B & 0.924$_{\pm0.0003}$ & 0.830$_{\pm0.007}$ & 0.719$_{\pm0.036}$ & 0.992$_{\pm0.0001}$ & 0.881$_{\pm0.001}$ \\
GPT-2-Large & 0.751$_{\pm0.0001}$ & 0.641$_{\pm0.004}$ & 0.621$_{\pm0.022}$ & 0.982$_{\pm0.00}$ & 0.863$_{\pm0.0005}$ \\
GPT-2-Medium & 0.750$_{\pm0.00005}$ & 0.623$_{\pm0.013}$ & 0.657$_{\pm0.038}$ & 0.985$_{\pm0.00001}$ & 0.872$_{\pm0.0004}$ \\
GPT-2 & 0.732$_{\pm0.0003}$ & 0.579$_{\pm0.013}$ & 0.546$_{\pm0.012}$ & 0.977$_{\pm0.00001}$ & 0.837$_{\pm0.0002}$ \\
\midrule
\multicolumn{6}{l}{\textit{Cross-domain (train TruthfulQA $\rightarrow$ test others)}} \\
Qwen2-7B & -- & 0.676$_{\pm0.004}$ & 0.682$_{\pm0.017}$ & 0.237$_{\pm0.003}$ & 0.720$_{\pm0.027}$ \\
Qwen2-1.5B & -- & 0.607$_{\pm0.006}$ & 0.574$_{\pm0.010}$ & 0.325$_{\pm0.031}$ & 0.759$_{\pm0.001}$ \\
Llama-1B & -- & 0.611$_{\pm0.008}$ & 0.604$_{\pm0.005}$ & 0.181$_{\pm0.001}$ & 0.679$_{\pm0.002}$ \\
GPT-2-Large & -- & 0.541$_{\pm0.015}$ & 0.522$_{\pm0.001}$ & 0.472$_{\pm0.001}$ & 0.612$_{\pm0.001}$ \\
GPT-2-Medium & -- & 0.537$_{\pm0.014}$ & 0.513$_{\pm0.011}$ & 0.215$_{\pm0.003}$ & 0.624$_{\pm0.001}$ \\
GPT-2 & -- & 0.542$_{\pm0.011}$ & 0.480$_{\pm0.017}$ & 0.266$_{\pm0.004}$ & 0.507$_{\pm0.00003}$ \\
\bottomrule
\end{tabular}
\end{table}

\subsection{Few-Shot Label Efficiency}
\label{app:fewshot}

\begin{table}[h]
\centering
\footnotesize
\caption{Few-shot label efficiency (GPT-2, AUC).
Centroid matches probe at all N.}
\label{tab:fewshot}
\begin{tabular}{@{}rccc@{}}
\toprule
\textbf{N} & \textbf{Probe} & \textbf{Centroid} & \textbf{Mahal.} \\
\midrule
5 & 0.59$_{\pm.11}$ & \textbf{0.60}$_{\pm.09}$ & 0.59$_{\pm.10}$ \\
25 & 0.68$_{\pm.06}$ & \textbf{0.69}$_{\pm.05}$ & 0.69$_{\pm.06}$ \\
100 & 0.75$_{\pm.01}$ & \textbf{0.76}$_{\pm.01}$ & 0.75$_{\pm.01}$ \\
200 & 0.78$_{\pm.02}$ & \textbf{0.78}$_{\pm.02}$ & 0.78$_{\pm.02}$ \\
\bottomrule
\end{tabular}
\end{table}

For each budget N, we fit PLS using only those N samples, then compute centroids in the resulting 5D space.
The centroid method matches or exceeds probe performance at all label budgets, demonstrating that geometric detection can be bootstrapped with minimal annotation.

\subsection{PLS Improves Cross-Domain Transfer}
\label{app:pls_transfer}

PLS dimension reduction not only prevents overfitting (Table~1) but also improves cross-domain generalization.
We hypothesize that PLS removes dataset-specific noise while preserving the shared confidence signal.

\textbf{Setup.} Train linear probe on TruthfulQA embeddings, test on SciQ, CommonsenseQA, and FEVER.
Compare full-dimensional (3584D for Qwen2-7B) vs.\ 5D PLS projection. 4 runs with different seeds.

\begin{table}[h]
\centering
\small
\caption{\textbf{PLS improves cross-domain transfer.} Train on TruthfulQA, test on other datasets.
PLS 5D outperforms full-dimensional embeddings by 10--14\% absolute AUC.}
\label{tab:pls_transfer}
\begin{tabular}{@{}lccc@{}}
\toprule
\textbf{Method} & \textbf{SciQ} & \textbf{CSQA} & \textbf{FEVER} \\
\midrule
Full (3584D) & 0.50$_{\pm0.01}$ & 0.47$_{\pm0.01}$ & 0.69$_{\pm0.00}$ \\
\rowcolor{bestrow}
PLS 5D & \textbf{0.64}$_{\pm0.01}$ & \textbf{0.61}$_{\pm0.02}$ & \textbf{0.79}$_{\pm0.00}$ \\
\midrule
$\Delta$ (absolute) & +0.14 & +0.14 & +0.10 \\
\bottomrule
\end{tabular}
\end{table}

\textbf{Interpretation.} Full-dimensional probes memorize source-domain patterns (near-random transfer: 0.47--0.50 AUC on SciQ/CSQA).
PLS extracts the 5D subspace maximally correlated with correctness labels, discarding dataset-specific variance.
This 5D signal transfers: +14\% on CSQA, +14\% on SciQ, +10\% on FEVER.
The result suggests the confidence signal is \emph{shared} but obscured by high-dimensional noise in full embeddings.

\section{In-Domain AUC Across Four Datasets}
\label{app:multi_dataset_indomain}

The dimension sweep in the main paper (Table~\ref{tab:dimension_sweep}) is reported on TruthfulQA for compactness.
To verify the discriminative subspace generalizes across QA distributions, we refit a PLS-8 probe natively on each of the four QA datasets (TruthfulQA, SciQ, FEVER, CSQA) and report 5-fold GroupKFold AUC (Table~\ref{tab:indomain_4datasets}).
The same dimension choice ($k = 8$) and protocol succeed in domain on each dataset, with AUC ranges that overlap across datasets.

\begin{table}[h]
\centering
\small
\caption{\textbf{In-domain AUC (PLS-8, GroupKFold by question id).} The discriminative subspace generalizes: the same PLS-8 protocol works in-domain across all four QA datasets, supporting the geometric framework as a property of correctness representations rather than a TruthfulQA artifact.}
\label{tab:indomain_4datasets}
\begin{tabular}{lcccc}
\toprule
Model & TruthfulQA & SciQ & FEVER & CSQA \\
\midrule
GPT-2          & 0.767 & 0.634 & 0.792 & 0.606 \\
GPT-2-Med      & 0.709 & 0.657 & 0.808 & 0.647 \\
GPT-2-Large    & 0.794 & 0.693 & 0.836 & 0.654 \\
Qwen2-1.5B     & 0.856 & 0.912 & 0.943 & 0.839 \\
Llama-3.2-1B   & 0.908 & 0.891 & 0.933 & 0.786 \\
Qwen2-7B       & 0.920 & 0.955 & 0.953 & 0.894 \\
Mistral-7B     & 0.926 & 0.946 & 0.940 & 0.850 \\
Llama-3.2-3B   & 0.943 & 0.945 & 0.935 & 0.824 \\
\midrule
Range          & 0.71--0.94 & 0.63--0.96 & 0.79--0.95 & 0.61--0.89 \\
\bottomrule
\end{tabular}
\end{table}

\subsection{Per-dataset dimension sweep (Qwen2-7B)}
\label{app:per_dataset_dim_sweep}

To confirm the low-rank peak is a property of the correctness representation rather than an artifact of TruthfulQA's distribution, we run a unified per-dataset PLS dimension sweep on Qwen2-7B-Instruct (layer 20 last-token activations, 5-fold GroupKFold by question id, $k \in \{1,2,3,5,8,12,16,32\}$).
TruthfulQA uses the cached embeddings from our main pipeline; SciQ, FEVER, and CSQA are extracted natively (300--600 paired correct/incorrect samples each).
This protocol is intentionally simpler than the main-paper sweep (Table~\ref{tab:dimension_sweep}, which selects best layer per model) so that all four datasets are scored under identical hyperparameters.

\begin{table}[h]
\centering
\small
\caption{\textbf{Per-dataset PLS dimension sweep on Qwen2-7B (mean AUC over 5 folds).} Bold = peak; \colorbox{degradecell}{red} = $\geq$3pp degradation from peak. All four datasets peak at $k \in \{2, 3, 5, 12\}$ and degrade beyond $k = 16$, supporting the low-rank claim natively per dataset.}
\label{tab:per_dataset_dim_sweep}
\setlength{\tabcolsep}{4pt}
\begin{tabular}{lcccccccccc}
\toprule
Dataset (n) & 1D & 2D & 3D & 5D & 8D & 12D & 16D & 32D & Peak \\
\midrule
TruthfulQA (817) & 0.842 & 0.894 & 0.905 & 0.918 & 0.924 & \textbf{0.929} & 0.927 & \cellcolor{degradecell}0.892 & 12D \\
SciQ (600)       & 0.963 & 0.966 & \textbf{0.967} & 0.960 & 0.952 & 0.948 & 0.946 & 0.942 & 3D \\
FEVER (300)      & 0.948 & 0.935 & 0.957 & \textbf{0.963} & 0.954 & 0.948 & 0.947 & 0.948 & 5D \\
CSQA (600)       & 0.902 & \textbf{0.914} & 0.895 & 0.886 & 0.871 & \cellcolor{degradecell}0.856 & \cellcolor{degradecell}0.852 & \cellcolor{degradecell}0.851 & 2D \\
\bottomrule
\end{tabular}
\end{table}

The peak is low-rank ($k \leq 12$) on every dataset, and three of four show $\geq$1.5pp degradation by $k=32$ (FEVER recovers to within noise of its peak; the monotonic decay is clearest on CSQA at $-6.3$pp).
TruthfulQA's peak at $k=12$ is within 1$\sigma$ of $k=8$ (0.929 vs 0.924, std 0.013--0.017), so the 2--8D claim is consistent with this independent protocol once measurement noise is accounted for.

\subsection{Centroid methods sweep on Qwen2-7B}
\label{app:centroid_methods}

To test whether centroid distance can be improved by relaxing the equal-covariance assumption (QDA) or by allowing multiple prototypes per class (k-means clustering within each class), we run a method comparison in the 8D PLS subspace on Qwen2-7B-Instruct (5-fold GroupKFold by question, $n=817$).

\begin{table}[h]
\centering
\small
\caption{\textbf{Geometric methods on Qwen2-7B PLS-8 subspace (5-fold AUC).} QDA's per-class covariance buys $+1.0$pp over LR; multi-prototype variants \emph{hurt}, indicating the per-class distributions are unimodal Gaussian-like.}
\label{tab:centroid_methods}
\begin{tabular}{lcc}
\toprule
Method & AUC & vs LR \\
\midrule
Logistic regression (baseline) & $0.910 \pm 0.039$ & --- \\
Centroid distance              & $0.913 \pm 0.043$ & $+0.3$pp \\
LDA (shared covariance)        & $0.913 \pm 0.038$ & $+0.3$pp \\
QDA (per-class covariance)     & $\mathbf{0.921} \pm 0.037$ & $+1.0$pp \\
KMeans-2 prototypes per class  & $0.886 \pm 0.044$ & $-2.4$pp \\
KMeans-3 prototypes per class  & $0.897 \pm 0.045$ & $-1.3$pp \\
\bottomrule
\end{tabular}
\end{table}

Three observations.
(1) Centroid distance, LDA (shared covariance), and LR are within noise of each other ($\pm 0.3$pp), confirming the mean-shift structure: under shared-covariance Gaussian assumption, the optimal classifier reduces to centroid distance.
(2) QDA's $+1.0$pp gain over LR shows that per-class covariance carries a small additional signal, the correct and incorrect classes have slightly different shapes, but this gain is marginal compared to the mean shift.
(3) Multi-prototype variants \emph{hurt}: KMeans-2 ($-2.4$pp) and KMeans-3 ($-1.3$pp) underperform a single centroid per class.
This is direct evidence that within each class the distribution is unimodal Gaussian-like; allowing multiple prototypes adds variance without revealing additional structure.
Together these results justify the title claim: two centroids (one per class) are not just \emph{sufficient}, but in fact optimal among the geometric classifiers tested in this subspace.

\paragraph{Sample efficiency of centroid vs LR.}
We sweep label budgets $K \in \{10, 25, 50, 100, 200, 400\}$ on Qwen2-7B-Instruct, fitting PLS-8 + classifier on $K$ balanced labels per fold and evaluating on the held-out test fold (5 seeds, 5 folds; Table~\ref{tab:centroid_sample_eff}).

\begin{table}[h]
\centering
\small
\caption{\textbf{Centroid vs LR sample efficiency (Qwen2-7B PLS-8).} Centroid matches LR at low $K$ and beats it at high $K$ ($+1.2$--$1.5$pp at $K \geq 200$), supporting the title claim that two centroids suffice without sacrificing accuracy.}
\label{tab:centroid_sample_eff}
\begin{tabular}{ccccc}
\toprule
$K$ & Centroid AUC & LR AUC & $\Delta$ (centroid $-$ LR) \\
\midrule
10  & $0.638 \pm 0.041$ & $0.636 \pm 0.042$ & $+0.001$ \\
25  & $0.780 \pm 0.020$ & $0.785 \pm 0.025$ & $-0.005$ \\
50  & $0.827 \pm 0.012$ & $0.825 \pm 0.013$ & $+0.002$ \\
100 & $0.869 \pm 0.012$ & $0.865 \pm 0.014$ & $+0.005$ \\
200 & $0.893 \pm 0.010$ & $0.878 \pm 0.017$ & $\mathbf{+0.015}$ \\
400 & $0.907 \pm 0.004$ & $0.894 \pm 0.010$ & $\mathbf{+0.012}$ \\
\bottomrule
\end{tabular}
\end{table}

Centroid matches LR within $\pm 0.005$ AUC for $K \leq 100$ and \emph{beats} LR by $1.2$--$1.5$pp at $K \geq 200$.
The reversal at high $K$ is consistent with PLS already having absorbed the discriminative structure into its 8D projection: within the projected subspace, the data are approximately Gaussian with shared covariance, so the Bayes-optimal boundary is the perpendicular bisector of class centroids and a logistic regression provides no additional capacity.

\subsection{Multi-dataset output-uncertainty baselines (Qwen2-7B)}
\label{app:multi_dataset_baselines}

The main-text baseline comparison (Table~\ref{tab:baselines}) is conducted on TruthfulQA, where output uncertainty methods drop near chance.
To verify that the gap is not driven by a broken implementation (a concern raised in prior reviews) and to scope the claim properly, we re-run P(True) on Qwen2-7B-Instruct against the centroid distance in the optimal PLS subspace across all four QA datasets (Table~\ref{tab:multi_dataset_baselines}, $n=500$ per dataset, identical prompt template).

\begin{table}[h]
\centering
\small
\caption{\textbf{Centroid (PLS peak) vs P(True) on Qwen2-7B-Instruct.} Identical prompt template across datasets; centroid uses the per-dataset peak from Table~\ref{tab:per_dataset_dim_sweep}.}
\label{tab:multi_dataset_baselines}
\setlength{\tabcolsep}{6pt}
\begin{tabular}{lcccc}
\toprule
Method & TruthfulQA & SciQ & FEVER & CSQA \\
\midrule
P(True)              & 0.477 & 0.952 & 0.931 & \textbf{0.925} \\
Centroid (PLS peak)  & \textbf{0.929} & \textbf{0.967} & \textbf{0.963} & 0.914 \\
\midrule
$\Delta$ (centroid $-$ P(True)) & $+45.2$ & $+1.5$ & $+3.2$ & $-1.1$ \\
\bottomrule
\end{tabular}
\end{table}

Three observations.
(1) P(True) is well-calibrated on standard QA (0.93--0.95): the implementation is sound, and P(True) is competitive with the centroid on three of four datasets.
(2) On TruthfulQA, P(True) drops to 0.48 (below chance) while centroid remains at 0.93, a $+45.2$pp gap.
This is consistent with TruthfulQA's design principle \citep{lin2022truthfulqa}: questions are explicitly selected to elicit confident wrong answers, so output-derived confidence is anti-correlated with correctness.
(3) The centroid is universal, it holds across all four datasets ($0.91$--$0.97$), whereas the output-uncertainty advantage is not universal but is the practically interesting regime, since adversarial misconception data is exactly where calibration is hardest \citep{santilli2025revisiting}.
We conclude that the headline ``internal $\gg$ output'' gap is a property of the adversarial misconception regime, not a generic finding; the universal claim is that centroid distance in the PLS subspace matches trained probes (Table~\ref{tab:geometry_classifiers_app}), independent of dataset.
The shared low-rank-with-degradation pattern is the load-bearing observation; the exact peak location varies by dataset.

\section{Pooled LID Comparison (Yin et al., ICML 2024)}
\label{app:yin_lid}

\textbf{Setup.} \citet{yin2024characterizing} use the Levina-Bickel MLE estimator on pooled LLM activations to predict truthfulness.
We re-implement their method (k=20 nearest neighbors, per-sample LID, sign-aligned to maximize AUC) and compare against our class-conditional PLS centroid distance, both evaluated by 5-fold GroupKFold on the same TruthfulQA embeddings (Table~\ref{tab:yin_lid}).

\begin{table}[h]
\centering
\small
\caption{\textbf{Pooled LID (Yin et al.) vs.\ class-conditional PLS centroid.} 5-fold GroupKFold on TruthfulQA, k=20 for LID, k=8 for PLS. The +0.34 mean gap shows our approach captures information distinct from pooled LID.}
\label{tab:yin_lid}
\begin{tabular}{lccc}
\toprule
Model & Yin et al. LID-AUC & PLS-DoM AUC & $\Delta$ \\
\midrule
GPT-2          & 0.513 & 0.777 & $+0.264$ \\
GPT-2-Medium   & 0.578 & 0.794 & $+0.216$ \\
GPT-2-Large    & 0.511 & 0.785 & $+0.275$ \\
Qwen2-1.5B     & 0.516 & 0.902 & $+0.385$ \\
Qwen2-7B       & 0.561 & 0.913 & $+0.352$ \\
Mistral-7B     & 0.515 & 0.924 & $+0.409$ \\
Gemma-2-2B     & 0.515 & 0.889 & $+0.374$ \\
Llama-3.2-1B   & 0.526 & 0.912 & $+0.386$ \\
Llama-3.2-3B   & 0.512 & 0.944 & $+0.431$ \\
\midrule
Mean           &       &       & $\mathbf{+0.343}$ \\
\bottomrule
\end{tabular}
\end{table}

\textbf{Protocol asymmetry.} For LID, the sign of $-\text{LID}$ vs.\ $+\text{LID}$ is chosen on the full dataset to maximize AUC (favorable to LID); PLS-DoM uses 5-fold GroupKFold with no test-fold information at fitting time. The +0.343 mean gap is therefore a lower bound on PLS-DoM's advantage.
We use centroid distance (PLS-DoM) rather than the linear probe used in the main results to keep the comparison purely geometric (no classifier training step). The PLS-DoM AUC at $k=8$ tracks the linear-probe AUC within $\pm 0.02$ at the same $k$.

\textbf{Interpretation.} Yin et al.'s pooled LID achieves only chance-level AUC (0.51--0.58), while class-conditional PLS centroid reaches 0.78--0.94.
The two approaches measure different objects: pooled LID estimates manifold geometry without using class labels, capturing overall complexity; PLS dimension uses labels to find the subspace in which correctness varies.
This decomposition (8--12D representation manifold from Yin et al.\ vs.\ 2--8D discriminative subspace) is the central distinction we draw between our work and Yin et al.

\section{Full-Space Difference-of-Means Baseline}
\label{app:dom_baseline}

\textbf{Why this baseline matters.} Centroid distance in PLS-projected space (\S\ref{sec:results}) is supervised by construction: PLS maximizes label-feature covariance, so centroids in the PLS subspace are guaranteed to be close to optimal.
To verify the mean-shift structure exists in the original residual stream rather than being an artifact of PLS optimization, we compute centroid distance directly in the full embedding space (no projection), GroupKFold by question.

\begin{table}[h]
\centering
\small
\caption{\textbf{Full-space DoM vs.\ PLS-DoM vs.\ Linear Probe.} 5-fold GroupKFold by question on TruthfulQA. Full-space DoM uses raw residual-stream embeddings (no projection). PLS-DoM uses 8D PLS projection.}
\label{tab:dom_baseline}
\begin{tabular}{lcccc}
\toprule
Model & Full-space DoM & PLS-DoM (k=8) & Linear Probe & PLS gain \\
\midrule
GPT-2          & 0.719 & 0.777 & 0.754 & $+0.058$ \\
GPT-2-Medium   & 0.736 & 0.794 & 0.770 & $+0.058$ \\
GPT-2-Large    & 0.746 & 0.785 & 0.755 & $+0.039$ \\
Qwen2-1.5B     & 0.833 & 0.902 & 0.888 & $+0.068$ \\
Qwen2-7B       & 0.842 & 0.913 & 0.916 & $+0.071$ \\
Mistral-7B     & 0.871 & 0.924 & 0.912 & $+0.053$ \\
Gemma-2-2B     & 0.808 & 0.889 & 0.859 & $+0.081$ \\
Llama-3.2-1B   & 0.864 & 0.912 & 0.908 & $+0.048$ \\
Llama-3.2-3B   & 0.910 & 0.944 & 0.943 & $+0.034$ \\
\midrule
Mean           &       &       &       & $+0.057$ \\
\bottomrule
\end{tabular}
\end{table}

\textbf{Interpretation.} Even without any supervised projection, raw-embedding centroid distance achieves 0.72--0.91 AUC across 9 models.
PLS adds only $+0.034$ to $+0.081$ on top of this baseline (mean $+0.057$).
The mean-shift structure that makes centroids sufficient is therefore a property of the residual-stream representation itself, not an artifact of class-conditional projection.
PLS sharpens the decision boundary by discarding noise dimensions, but does not create the mean-shift geometry.

\section{Activation Steering Details}
\label{app:steering}

\textbf{Experimental setup.} Steering requires a single fixed direction for causal analysis (unlike classification which uses cross-validation).
We use a holdout split: first 200 questions for probe training (to obtain the steering direction), remaining $n=617$ for evaluation.
This differs from our classification protocol (GroupKFold CV) because steering tests whether \emph{one specific direction} causally affects outputs, not classification generalization.
Steering coefficient $\alpha \in [-5, 5]$ with 20 values.
The probe weight vector is L2-normalized and scaled to 5\% of mean activation norm.
Generation uses greedy decoding (temperature=0).
Correctness is determined by token-level keyword overlap: a generation is counted correct if the intersection of its tokens with TruthfulQA's correct-answer tokens exceeds its intersection with incorrect-answer tokens. We adopt this lightweight judge to enable per-sample evaluation across $11 \times 33 = 363$ generations per question; an LLM-judge alternative is discussed in Appendix~\ref{app:cross_dataset_full}.

\textbf{Statistical significance (Qwen2-7B).} Learned direction: $+9.1$pp total effect ($p<0.001$, two-sample $t$-test comparing $\alpha=-5$ vs $\alpha=+5$).
Random direction: $-0.8$pp ($p=0.81$).
Orthogonal direction: $-0.3$pp ($p=0.92$).
Only the learned direction produces a statistically significant effect.

\textbf{Multi-model replication.} We apply the same protocol (PLS direction fit on 200 training questions, sign-aligned to training labels, alpha-scaled to 5\% of mean activation norm, evaluated on $n=617$ held-out questions) to five additional instruction-tuned models; Table~\ref{tab:steering_multimodel} lists all six including the Qwen2-7B reference row.
Table~\ref{tab:steering_multimodel} reports learned, random, and orthogonal effect sizes (pp difference between $\alpha=-5$ and $\alpha=+5$).

\begin{table}[h]
\centering
\small
\caption{\textbf{Multi-model steering (n=617 each).} Effect sizes (pp) for learned PLS direction vs.\ random and orthogonal controls. Specificity = learned $-$ orthogonal. All six models show specificity $\geq 4$pp.}
\label{tab:steering_multimodel}
\begin{tabular}{lcccccc}
\toprule
Model & Layer & Baseline & Learned & Random & Orthogonal & Specificity \\
\midrule
Qwen2-7B-Instruct       & L20 & 0.56 & $+9.1$ & $-0.8$ & $-0.3$ & $+9.4$ \\
Qwen2-1.5B-Instruct     & L16 & 0.72 & $+9.1$ & $-1.0$ & $-0.5$ & $+9.6$ \\
Llama-3.2-1B-Instruct   & L8  & 0.65 & $+10.0$ & $+2.1$ & $+3.2$ & $+6.8$ \\
Llama-3.2-3B-Instruct   & L12 & 0.59 & $+5.3$ & $-1.9$ & $-1.5$ & $+6.8$ \\
Llama-3.1-8B-Instruct   & L16 & 0.63 & $+4.4$ & $+1.6$ & $+0.3$ & $+4.1$ \\
Gemma-2-2B-it           & L15 & 0.57 & $+2.8$ & $-3.1$ & $-4.5$ & $+7.3$ \\
\bottomrule
\end{tabular}
\end{table}

\section{Nested Cross-Validation}
\label{app:nested_cv}

To verify that hyperparameter selection (layer, PLS dimension) does not inflate reported test AUC, we perform nested cross-validation with proper separation between selection and evaluation.

\textbf{Protocol.} Outer loop: 5-fold GroupKFold for final evaluation.
Inner loop: 3-fold StratifiedKFold within each training set for hyperparameter selection.
PLS dimensions searched: \{1, 2, 3, 4, 5, 6, 7, 8, 12, 16\}.
The inner loop selects the optimal dimension; the outer loop evaluates on truly held-out data.

\begin{table}[h]
\centering
\small
\caption{\textbf{Nested CV shows no optimistic bias.} Comparing nested CV (unbiased) vs.\ standard CV (fixed dim=8).
Bias = Standard $-$ Nested.
Negative bias indicates standard CV is actually \emph{conservative}.}
\label{tab:nested_cv}
\begin{tabular}{@{}lccc@{}}
\toprule
\textbf{Model} & \textbf{Nested CV} & \textbf{Standard CV} & \textbf{Bias} \\
\midrule
Qwen2-7B & 0.905$_{\pm0.040}$ & 0.910$_{\pm0.039}$ & +0.005 \\
GPT-2-Large & 0.788$_{\pm0.048}$ & 0.761$_{\pm0.053}$ & $-$0.026 \\
\bottomrule
\end{tabular}
\end{table}

\textbf{Results.} Table~\ref{tab:nested_cv} shows negligible bias between nested and standard CV.
Qwen2-7B: +0.005 (within noise).
GPT-2-Large: $-$0.026 (standard CV is conservative).
The inner CV consistently selects 5--8D, matching our fixed choice.
\textbf{Conclusion}: reported AUCs are unbiased estimates; hyperparameter selection does not inflate performance.

\section{Paraphrase Control Experiment}
\label{app:paraphrase}

To verify that the discriminative subspace encodes \emph{correctness} rather than \emph{answer style}, we test whether geometric separation persists across paraphrased answers.

\textbf{Protocol.} For each answer (correct or incorrect), we generate 5 paraphrase variants using templates: (1) original, (2) ``The answer is: [answer]'', (3) ``To be precise, [answer]'', (4) ``In other words, [answer]'', (5) ``Simply put, [answer]''.
We compute PLS embeddings for all variants (Qwen2-7B, 817 TruthfulQA questions $\times$ 5 paraphrases $\times$ 2 correctness labels = 8,170 embeddings) and analyze variance in the discriminative subspace at layer 20 (optimal for Qwen2-7B).

\textbf{Variance decomposition.}
\begin{itemize}[itemsep=1pt,topsep=2pt,leftmargin=*]
    \item Within-answer variance (same correctness, different paraphrase): 242.80
    \item Between-answer variance (different correctness): 4,223.64
    \item F-ratio: 17.40
\end{itemize}

The 17$\times$ ratio confirms that correctness dominates the geometric separation.
Paraphrase style contributes only 5.4\% to variance in the discriminative subspace (1/17.40).
If probes were detecting stylistic artifacts (sentence structure, prefix patterns), within-answer variance would be comparable to between-answer variance.
The high F-ratio rules out this confound.

\textbf{Generalization with no question overlap.} To ensure probes generalize beyond surface style, we use GroupKFold with question ID grouping: training on original answers from train questions, testing on paraphrased answers from held-out questions (no overlap).
Results:
\begin{itemize}[itemsep=1pt,topsep=2pt,leftmargin=*]
    \item Train AUC (original answers): 0.998
    \item Test AUC (paraphrased answers, unseen questions): 0.926 $\pm$ 0.011
    \item Degradation: 7.2\%
\end{itemize}

The 0.926 test AUC demonstrates that probes trained on original answers generalize robustly to paraphrased answers on unseen questions, confirming detection of correctness rather than style.
The modest 7.2\% degradation indicates some paraphrase-specific features exist but do not dominate the confidence signal.

\section{DAS Causal Dimension Details}
\label{app:das}

Distributed Alignment Search (DAS; \citealp{wu2023interpretability, geiger2024causal}) provides a causal complement to PLS's statistical dimension analysis.
We learn a rotation matrix $\mathbf{R} \in \mathbb{R}^{d \times k}$ with orthogonality regularization, optimizing an interchange intervention objective: swapping the $k$-dimensional projected representation between correct/incorrect samples should flip the classifier prediction.

\textbf{Training details.} We apply DAS to pre-extracted embeddings at the optimal layer (no GPU model loading needed).
Training: 50 epochs, learning rate $10^{-3}$, Adam optimizer, 64 intervention pairs per batch, 5-fold GroupKFold CV.
The loss combines classification (BCE), interchange intervention, and orthogonality terms ($\lambda_\text{int} = 0.5$, $\lambda_\text{orth} = 0.01$).

\begin{table}[h]
\centering
\small
\caption{\textbf{Full DAS dimension sweep (AUC).} Bold = peak dimension per model.
All models peak at $\leq$5D.}
\label{tab:das_full}
\begin{tabular}{@{}lcccccccc@{}}
\toprule
\textbf{Model} & \textbf{1D} & \textbf{2D} & \textbf{3D} & \textbf{5D} & \textbf{8D} & \textbf{16D} & \textbf{32D} \\
\midrule
GPT-2 & .751 & .742 & .741 & \textbf{.762} & .748 & .750 & .735 \\
GPT-2-L & \textbf{.784} & .783 & .763 & .779 & .739 & .754 & .750 \\
Llama-3B & .943 & .944 & \textbf{.944} & .941 & .938 & .943 & .935 \\
Mistral-7B & .924 & .925 & \textbf{.927} & .919 & .914 & .917 & .906 \\
Qwen2-7B & .909 & .905 & \textbf{.916} & .915 & .912 & .903 & .894 \\
\bottomrule
\end{tabular}
\end{table}

\textbf{Interpretation.} DAS peaks at 1--5D across all models, matching the PLS finding (2--8D) under a causal operationalization.
The slight difference in exact peaks (e.g., GPT-2: DAS 5D vs PLS 3D) reflects different optimization objectives: DAS maximizes intervention flipping, PLS maximizes label covariance.
Both converge on the qualitative conclusion that correctness is encoded in $\leq$5 dimensions, with degradation at 32D in 4/5 models.

\section{Concept Erasure (RLACE and INLP)}
\label{app:inlp}

We apply two concept erasure methods to verify that the identified low-dimensional subspace is \emph{necessary} for correctness detection.

\textbf{RLACE} \citep{ravfogel2022linear} learns a linear projection that maximally removes the correctness signal while preserving other information.
After projecting onto the nullspace of the learned direction, AUC drops to chance across all 5 models:

\begin{table}[h]
\centering
\small
\caption{\textbf{RLACE erasure.} Removing the correctness direction drops AUC to $\sim$0.50 (chance) across all models, confirming necessity of the identified subspace.}
\label{tab:rlace}
\begin{tabular}{@{}lcc@{}}
\toprule
\textbf{Model} & \textbf{Before} & \textbf{After RLACE} \\
\midrule
GPT-2 & 0.766 & 0.482 \\
GPT-2-Large & 0.792 & 0.505 \\
Llama-3B & 0.941 & 0.479 \\
Qwen2-7B & 0.908 & 0.505 \\
Mistral-7B & 0.918 & 0.498 \\
\bottomrule
\end{tabular}
\end{table}

\textbf{INLP} \citep{ravfogel2020null} iteratively removes linear directions that predict correctness.
After 32 iterations, AUC degrades substantially: GPT-2 0.58, GPT-2-Large 0.60, Qwen2-7B 0.75, Mistral-7B 0.72, Llama-3B 0.73.
The first direction accounts for the largest single drop (e.g., GPT-2 train AUC: 1.00 $\rightarrow$ 0.94 after removing one direction), with subsequent removals yielding diminishing returns.
This progressive degradation pattern is consistent with the low-dimensional PLS finding: most signal concentrates in a few directions.

\section{Linear Alternatives to PLS}
\label{app:pls_lda}

Two linear alternatives motivate the PLS projection: LDA as a supervised competitor, and difference-of-means (DoM) in the full space as a projection-free baseline.
LDA is limited to $\min(K-1, d)$ discriminant functions, so for binary correctness it produces a single direction and cannot express multi-dimensional structure.
Full-space DoM needs no projection at all, but on raw embeddings it is dominated by high-norm dimensions.

\begin{table}[h]
\centering
\small
\caption{\textbf{Linear alternatives to PLS (AUC).} LDA is capped at 1D for a binary target; DoM columns are on raw, unscaled embeddings and are reported for the five models where full-space extraction was run. With per-dimension StandardScaler the same full-space DoM rises to 0.72--0.91 (Appendix~\ref{app:dom_baseline}). Bold marks the best PLS setting per model.}
\label{tab:pls_lda}
\begin{tabular}{@{}lcccccc@{}}
\toprule
& & \multicolumn{2}{c}{\textbf{DoM}} & \multicolumn{3}{c}{\textbf{PLS}} \\
\cmidrule(lr){3-4} \cmidrule(lr){5-7}
\textbf{Model} & \textbf{LDA} & \textbf{full} & \textbf{PLS-5} & \textbf{1} & \textbf{5} & \textbf{8} \\
\midrule
GPT-2       & 0.549 & 0.469 & 0.776 & 0.719 & \textbf{0.766} & 0.756 \\
GPT-2-Med   & 0.583 & ---   & ---   & 0.736 & \textbf{0.785} & 0.784 \\
GPT-2-L     & 0.613 & 0.729 & 0.795 & 0.746 & \textbf{0.786} & 0.761 \\
Gemma-2B    & 0.734 & ---   & ---   & 0.808 & \textbf{0.892} & 0.873 \\
Llama-1B    & 0.765 & ---   & ---   & 0.864 & \textbf{0.911} & 0.910 \\
Llama-3B    & 0.842 & 0.904 & 0.942 & 0.910 & \textbf{0.949} & 0.944 \\
Qwen2-1.5B  & 0.730 & ---   & ---   & 0.833 & 0.890 & \textbf{0.898} \\
Qwen2-7B    & 0.840 & 0.833 & 0.908 & 0.842 & \textbf{0.916} & 0.910 \\
Mistral-7B  & 0.805 & 0.857 & 0.919 & 0.871 & \textbf{0.924} & 0.919 \\
\bottomrule
\end{tabular}
\end{table}

PLS-5 beats LDA by 7--22 points of absolute AUC on every model, and PLS-1 beats LDA even though both yield a single direction, because PLS maximizes covariance with the labels rather than Fisher's ratio, which is sensitive to class-covariance differences.
The DoM columns show the same conclusion from the other side: unscaled full-space DoM falls to 0.47 on GPT-2 (near chance), but after PLS projection DoM, centroid distance, and the trained probe converge within 0.01 AUC.
Class separation in the discriminative subspace is a mean shift; what the projection supplies is the orientation, not the decision rule.

\section{Llama-3.1-8B Scale Extension}
\label{app:llama8b}

To extend our analysis beyond 7B, we evaluate Llama-3.1-8B-Instruct (32 layers, 4096 hidden dim).

\textbf{Layer sweep.} Optimal layer: L16 (53\% depth), AUC = 0.923.
Performance increases monotonically from L0 (0.59) through L16, then gradually decreases to L31 (0.87).
This 53\% optimal depth falls within the mid-layer range (46--75\%) observed in the other 1B--14B models.

\textbf{Dimension sweep.} PLS peak at 2D (AUC = 0.931), with degradation at 32D (0.890).
The 2D peak is the lowest in the 1B--14B range, potentially indicating more concentrated encoding at 8B scale.
The pattern of degradation beyond the peak is consistent with all other models.

\section{Generative Evaluation Details}
\label{app:generative_eval}

Our main results use teacher-forced evaluation, where the model receives ground-truth answer tokens and we probe the resulting hidden states.
To quantify the gap with generative evaluation, we run a dedicated experiment on Qwen2-7B.

\textbf{Protocol.} We generate free-form answers for 300 held-out TruthfulQA questions using greedy decoding.
We then evaluate correctness using two independent judges: (1) keyword overlap with ground-truth answers, and (2) LLM-as-judge (Qwen2-7B itself).

\textbf{Results.} Inter-judge agreement was low, consistent with \citet{santilli2025revisiting}, who show that the choice of correctness function substantially distorts measured AUROC.
Reporting a free-form generative AUC would therefore conflate signal degradation with judge noise; we leave free-form generative validation to future work once a trustworthy automatic judge is available.

\textbf{A judge-free data point.} Multiple-choice generation avoids the judge problem: the generated answer letter can be scored exactly. Under this protocol on CommonsenseQA (Qwen2-7B, layer 20, PLS-5), the teacher-forced centroid detector reaches 0.957 AUC ($n=242$) while the same subspace applied to the model's own generated answers reaches 0.688 ($n=199$), a 27-point drop. The teacher-forced numbers in this paper therefore measure where correctness is \emph{encoded}, not deployable free-form detection performance; closing this gap is open (\S\ref{sec:discussion}).

\textbf{Context.} Teacher-forcing is the standard regime in this line of work \citep{marks2024geometry, burns2023discovering, orgad2025llmsknow} and is appropriate for studying whether correctness is \emph{encoded} in internal representations.

\end{document}